\documentclass{article}

\usepackage{arxiv}
\usepackage{tabularx}
\usepackage[ruled,linesnumbered,commentsnumbered]{algorithm2e}
\usepackage[utf8]{inputenc}
\usepackage{graphicx}
\usepackage[T1]{fontenc}
\usepackage{mathptmx}
\usepackage{hyperref}
\usepackage{tablefootnote}
\usepackage[nottoc]{tocbibind}
\usepackage{caption}
\usepackage{subcaption} 
\usepackage{footnote}
\usepackage{amsmath}
\usepackage{amssymb}
\usepackage{mathrsfs}
\usepackage{array}
\usepackage{dcolumn}
\usepackage{bm}
\usepackage[utf8]{inputenc} 
\usepackage[T1]{fontenc}    
\usepackage{hyperref}       
\usepackage{url}            
\usepackage{booktabs}       
\usepackage{amsfonts}       
\usepackage{nicefrac}       
\usepackage{microtype}      
\usepackage{lipsum}

\usepackage[utf8]{inputenc}
\title{ChaosNet: A Chaos based Artificial Neural Network Architecture for Classification}

\author{
Harikrishnan Nellippallil Balakrishnan$^{*}$, Aditi Kathpalia$^{*}$, Snehanshu Saha$^{**}$, Nithin Nagaraj$^{*}$\\
 $^{*}$Consciousness Studies Programme,\\ National Institute of Advanced Studies, Indian Institute of Science Campus, Bengaluru, India. \\
$^{**}$Center for AstroInformatics,\\ Modeling and Simulation (CAMS) \& Dept. of Computer Science, PES University, Bengaluru, India.\\
  \texttt{harikrishnan.nb@nias.res.in, kathpaliaaditi@nias.res.in, snehanshusaha@pes.edu, nithin@nias.res.in  } \\
}
\begin{document}
\maketitle

\begin{abstract}
Inspired by chaotic firing of neurons in the brain, we propose \verb+ChaosNet+ -- a novel chaos based artificial neural network architecture for classification tasks. \verb+ChaosNet+ is built using layers of neurons, each of which is a 1D chaotic map known as the Generalized Lur\"{o}th Series (GLS) which has been shown in earlier works to possess very useful properties for compression, cryptography and for computing  XOR and other logical operations. In this work, we design a novel learning algorithm on \verb+ChaosNet+ that exploits the topological transitivity property of the chaotic GLS neurons. The proposed learning algorithm gives consistently good performance accuracy in a number of classification tasks on well known publicly available datasets with very limited training samples. Even with as low as $7$ (or fewer) training samples/class (which accounts for less than 0.05\% of the total available data), \verb+ChaosNet+ yields performance accuracies in the range $73.89 \% - 98.33 \%$. We demonstrate the robustness of \verb+ChaosNet+ to additive parameter noise and also provide an example implementation of a 2-layer \verb+ChaosNet+ for enhancing classification accuracy. We envisage the development of several other novel learning algorithms on \verb+ChaosNet+ in the near future.
\end{abstract}

{\it Chaos has been empirically found in the brain at several spatio-temporal scales\cite{therechaos1, therechaos2}. In fact, individual neurons in the brain are known to exhibit chaotic bursting activity and several neuronal models such as the Hindmarsh-Rose neuron model exhibit complex chaotic dynamics~\cite{fan1993bifurcations}. Though Artificial Neural Networks (ANN) such as Recurrent Neural Networks exhibit chaos, to our knowledge, there have been no successful attempts in building an ANN for classification tasks which is entirely comprised of neurons which are individually chaotic. Building on our earlier research, in this work, we propose  \verb+ChaosNet+ -- an ANN built out of neurons -- each of which is a 1D chaotic map known as Generalized Lur\"{o}th Series (GLS). GLS has been shown to have salient properties such as ability to encode and decode information losslessly with Shannon optimality, computing logical operations (XOR, AND etc.), universal approximation property and ergodicity (mixing) for cryptography applications. In this work, \verb+ChaosNet+ exploits the topological transitivity property of chaotic GLS neurons for classification tasks with state-of-the art accuracies in the low training sample regime. This work, inspired by the chaotic nature of neurons in the brain, demonstrates the unreasonable effectiveness of chaos and its properties for machine learning. It also paves the way for designing and implementing other novel learning algorithms on the \verb+ChaosNet+ architecture.}
\section{\label{sec:Intro} Introduction}

With the success of Artificial Intelligence, learning through algorithms such as Machine Learning (ML) and Deep Learning (DL) has become an area of intense activity and popularity with applications reaching almost every field known to humanity. These include -- medical diagnosis~\cite{ding2018deep}, computer vision, cyber-security~\cite{harikrishnan2018machine}, natural language processing, speech processing\cite{speechrnn}, just to name a few. These algorithms, though inspired by the biological brain, are remotely related to the biological process of learning and memory encoding. The learning procedures used in these artificial neural networks (ANNs) to modify weights and biases are based on optimization techniques and minimization of loss/error functions. The ANNs at present use an enormous number of hyperparamters which are fixed by an ad-hoc procedure for improving prediction as more and more new data is input into the system. These synaptic changes employed are solely data-driven and have little or no rigorous theoretical
backing~\cite{saxe2018information, tishby2015deep}. Furthermore, for accurate prediction/classification, these methods require enormous amount of training data that captures the distribution of the target classes.

Despite their tremendous success, ANNs are nowhere close to the human mind for accomplishing tasks such as natural language processing. To incorporate the excellent learning abilities of the human brain, as well as, to understand the brain better, researchers are now focusing on developing biologically inspired algorithms and architectures. This is being done both in the context of learning~\cite{delahunt2018putting} and memory encoding~\cite{aihara1990chaotic, crook2001novel, freeman1975mass, chang1996parameter, kozma1999possible, tsuda1992dynamic, nicolis1985chaotic, kaneko1986lyapunov, kaneko1990clustering, kathpalia2019novel, aram2017using}.

One of the most interesting properties of the brain is its ability to exhibit {\it Chaos}\cite{alligood1996chaos} -- the phenomenon of complex unpredictable and random-like behaviour arising from simple deterministic nonlinear systems\footnote{Deterministic chaos is characterized by the {\it Butterfly Effect --} sensitive dependence of behaviour to minute changes in initial conditions.}. The dynamics in electroencephalogram (EEG) signals is known to be chaotic~\cite{therechaos2}. The sensitivity to small shifts in internal functional parameters of a neuronal system helps to get desired response to different influences. This attribute resembles the dynamical properties of chaotic systems~\cite{babloyantz1996brain, barras2013mind, elbert1995chaotic}. Moreover, it is seen that the brain may not reach a state of equilibrium after a transient, but is constantly alternating between different states. For this reason, it is suggested that with the change in functional parameters of the neurons, the brain is able to exhibit different behaviours -- periodic orbits, weak chaos and strong chaos for different purposes. For example, there has been evidence to suggest that  weak chaos may be good for learning~\cite{sprott2013chaos} and periodic activity in the brain being useful for attention related tasks~\cite{baghdadi2015chaotic}. Thus, chaotic regimes exhibiting a wide variety of behaviors help the brain in quick adaptation to changing conditions. 

Chaotic behaviour is exhibited not only by brain networks which are composed of billions of neurons, but the dynamics at the neuronal level (cellular and sub-cellular) are also chaotic~\cite{therechaos2}. Impulse trains produced by these neurons are actually responsible for the transmission and storage of information in the brain. These impulses or \emph{action potentials} are generated when different ions cross the axonal membrane causing a change in the voltage across it.  Hodgkin and Huxley were the first to propose a dynamical system's model for the interaction between the ion channels and axon membrane, that is capable of generating realistic action potentials~\cite{hodgkin}. Later, its simplified versions such as the Hindmarsh-Rose model~\cite{hindmarsh1984model} and the Fitzugh-Nagumo~\cite{fitzhugh1961impulses, nagumo1962active} model were proposed. All these models exhibit chaotic behaviour.

Although there exist some artificial neural networks which display chaotic dynamics (an example is Recurrent Neural Networks~\cite{chaotic_rnn}), to the best of our knowledge, none of the architectures proposed for classification tasks till date exhibit chaos at the level of individual neurons. However, for a theoretical explanation of memory encoding in the brain, many chaotic neuron models have been suggested. These include the Aihara model~\cite{aihara1990chaotic} which has been utilized for memory encoding in unstable periodic orbits of the network~\cite{crook2001novel}. Freeman, Kozma and group have developed chaotic models inspired from the mammalian olfactory network to explain the process of memorizing of odors~\cite{freeman1975mass, chang1996parameter, kozma1999possible}. Chaotic neural networks have been studied also by Tsuda et al. for their functional roles as short term memory generators as well a dynamic link for long term memory~\cite{tsuda1992dynamic, nicolis1985chaotic}. Kaneko has explored the dynamical properties of globally coupled chaotic maps  suggesting possible biological information processing capabilities of these networks~\cite{kaneko1986lyapunov, kaneko1990clustering}. Our group (two of the authors) has also proposed a biologically-inspired network architecture with chaotic neurons which is capable of memory encoding~\cite{kathpalia2019novel}.

In this work, we propose \verb+ChaosNet+ -- an ANN built out of 1D chaotic map Generalized Lur\"{o}th Series (GLS) as its individual neurons. This network can accomplish classification tasks by learning with limited training samples. \verb+ChaosNet+ is developed as an attempt to use some of the best properties of biological neural networks arising as a result of rich chaotic behavior of individual neurons and is shown to accomplish challenging classification tasks comparable to or better than conventional ANNs while requiring far less training samples.

Choice of 1D maps as neurons in \verb+ChaosNet+ helps to keep the processing simple and at the same time exploit the useful properties of chaos. Our group (two of the authors) has discussed the use of the property of topological transitivity of these GLS neurons for classification~\cite{nagaraj2019novel}. Building on this initial work, in the current manuscript, we propose a novel and improved version of topological transitivity scheme for classification. This improved novel scheme, proposed for the very first time in this paper, utilizes a `spike-count rate' like property of the firing of chaotic neurons as a neural code for learning and is inspired from biological neurons. Moreover, the network is capable of displaying hierarchical architecture which can integrate information as it is passed on to higher levels (deeper layers in the network). The proposed classification scheme is rigorously tested on publicly available datasets -- MNIST, KDDCup'99, Exoplanet and Iris.

Current state-of-the-art algorithms in the AI community rely heavily on availability of enormous amounts of training data. However, there are several practical scenarios where huge amounts of training data may not be available~\cite{distributionkddcup}.Learning from limited samples plays a key role especially in a field like cyber security where new malware attacks occur frequently. For example, detecting zero day malware attack demands the algorithms to learn from fewer data samples.  Our proposed  \verb+ChaosNet+ architecture addresses this issue. 
The paper is organized as follows. GLS-neuron and its properties are described in section II. In section III, we introduce $1$-layer \verb+ChaosNet+ architecture and its application for topological transitivity symbolic sequence based classification algorithm. Experiments, results and discussion including parameter noise analysis of the $1$-layer TT-SS classification algorithm are dealt in section IV. Multilayer \verb+ChaosNet+ architecture is introduced in section V. We conclude with future research direction in section VI.

\section{\label{sec:GLS_Neuron_and_props} GLS-Neuron and its properties}
The neuron we propose is a piece-wise linear 1D chaotic map known as Generalized Lur\"{o}th Series or GLS~\cite{dajani2002ergodic}. The well known Tent map, Binary map and their skewed cousins are all examples of GLS. Mathematically, the types of GLS neurons we use in this work are described below. 
\subsection{\label{sec:GLS} GLS-Neuron types: $T_{Skew-Tent}(x)$ and $T_{Skew-Binary}(x)$}
\begin{figure}[!h]
	\centering
		\includegraphics[scale=0.25]{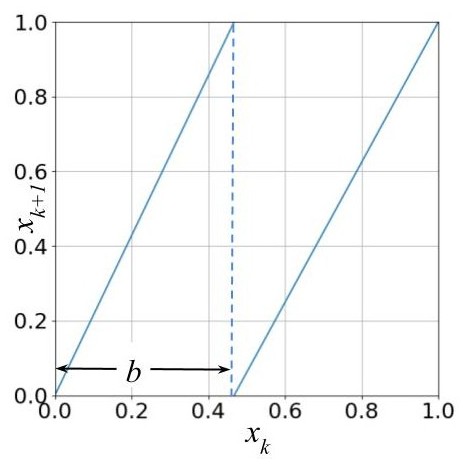}
		\includegraphics[scale=0.25]{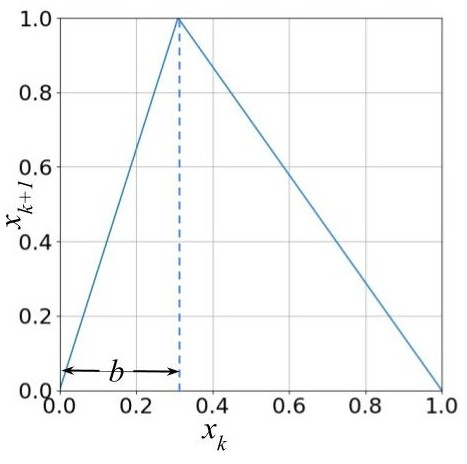}\\
		~~~(a)~~~~~~~~~~~~~~~~~~~~~~~~~~~~~~~~~~~~~~~~~~~~~~~(b)
		\caption{Generalized Lur\"{o}th Series, GLS-Neuron, is fundamentally of two types: (a) Left: Skew-Binary map $T_{Skew-Binary}$. (b) Right: Skew-Tent map $T_{Skew-Tent}.$ }\label{fig_GLSmaps}
\end{figure}

$T_{Skew-Binary}: [0,1) \rightarrow [0,1)$ is defined as: 
\begin{eqnarray*}
T_{Skew-Binary}(x)  =  \left\{\begin{matrix}
\frac{x}{b}&, ~~~~ 0 \leq x < b, \\ 
\frac{(x-b)}{(1 - b)}&, ~~~~ b \leq x < 1, 
\end{matrix}\right. \\
\end{eqnarray*}
where $x \in [0,1)$ and $0 < b <1$. Refer to Figure~\ref{fig_GLSmaps}(a). 
\\
$T_{Skew-Tent}: [0,1) \rightarrow [0,1)$ is defined as: 
\begin{eqnarray*}
T_{Skew-Tent}(x)  =  \left\{\begin{matrix}
\frac{x}{b}&, ~~~~ 0 \leq x < b, \\ 
\frac{(1-x)}{(1 - b)}&, ~~~~ b \leq x < 1, 
\end{matrix}\right. \\
\end{eqnarray*}

where $x \in [0,1)$ and $0 < b <1$. Refer to Figure~\ref{fig_GLSmaps}(b).\\ 
%

The symbols {\bf L (or 0)} and {\bf R (or 1)} are associated with the intervals $[0,b)$ and $(b,1)$ respectively, thereby defining the \emph{symbolic sequence}~\footnote{Let $X = \{x_0, x_1, x_2, \ldots \}$ be the trajectory of a chaotic map with initial condition $x_0$, where $x_i \in [U, V)$. The interval $[U, V)$ is partitioned into $k$ sub intervals denoted as $I_0, I_1, \ldots, I_{k-1}$. If $x_i \in I_j$ then we denote $x_i$ by the symbol $j \in \{0, 1, \ldots, k-1\}$. The new sequence of symbol $\{j_0, j_1, \ldots, j_{k-1\}}$ is the symbolic sequence of the trajectory of $X$.} for every trajectory starting from an initial value on the map.  We enumerate some salient properties of the GLS-neuron:
\begin{enumerate}
\item Each GLS-neuron has two parameters - an initial activity value $x_0$ and the skew value $b$. The parameter $b$ defines the \emph{generating Markov partition}\footnote{Generating Markov Partition or GMP is based on splitting the state space into a complete set of disjoint regions, namely, it covers all state space and enables associating a one-to-one correspondence between trajectories and itinerary sequences of symbols (L and R) without losing any information\cite{GMP}.} for the map and also acts as \emph{internal discrimination threshold} of the neuron which will be used for feature extraction. 

\item GLS-neuron can fire either chaotically or in a periodic fashion (of any finite period length) depending on the value of the initial activity value $x_0$. The degree of chaos is controlled by the skew parameter $b$. The lyapunov exponent of the map\cite{nagaraj2008novel} is given by $\lambda_{b} = -b\ln(b) -(1-b)\ln(1-b)$. If the base of the logarithm is chosen as 2, then $\lambda_b = H(S_{x_0})$ (bits/iteration) where $S_{x_0}$ is the symbolic sequence of the trajectory obtained by iterating the initial neural activity $x_0$ and $H(\cdot)$ is Shannon Entropy in bits/symbol. For $0 < b < 1$, $\lambda_b > 0$.

\item Any finite length input stream of bits can be \emph{losslessly} compressed as an initial activity value $x_0$ on the neuron by performing a backward iteration on the map. Further, such a lossless compression scheme has been previously shown to be \emph{Shannon optimal}\cite{nagaraj2009arithmetic}.

\item Error detection properties can be incorporated into GLS-neuron to counter the effect of noise\cite{nagaraj2019cantor}. 

\item Owing to its excellent chaotic and ergodic (mixing) properties, GLS (and other related 1D maps) has been employed in cryptography~\cite{nagaraj2009arithmetic, nagaraj2008novel, wong2010simultaneous}. 

\item Recently, two of the authors of the present work have proposed a compression-based neural architecture for memory encoding and decoding using GLS~\cite{kathpalia2019novel}.

\item GLS-neuron can compute logical operations such as XOR, AND, NOT, OR etc. by switching between appropriate maps as described in a previous work\cite{nagaraj2008novel, kathpalia2019novel}. 

\item GLS-neuron has the property of \emph{topological transitivity} - defined in a later section, which we will employ to perform classification. 

\item A single layer of a finite number of GLS-neurons satisfies a version of the \emph{Universal Approximation Theorem} which we shall prove in a subsequent section of this paper.
\end{enumerate}

The aforementioned properties make GLS-neurons an ideal choice for building our novel architecture for classification.

\section{\label{sec:Chaosnet} ChaosNet: The Proposed Architecture}

In this section, we first introduce the novel \verb+ChaosNet+ architecture followed by a description of the single-layer Topological Transitivity-Symbolic Sequence classification algorithm. We discuss the key principles behind this algorithm, its parameters and hyperparameters, an illustrative example and a proof of the Universal Approximation Theorem.

\subsection{Single layer ChaosNet Topological Transitivity - Symbolic Sequence (TT-SS) based Classification Algorithm}
\begin{figure}[!h]
    \centerline{ \includegraphics[width=0.6\textwidth]{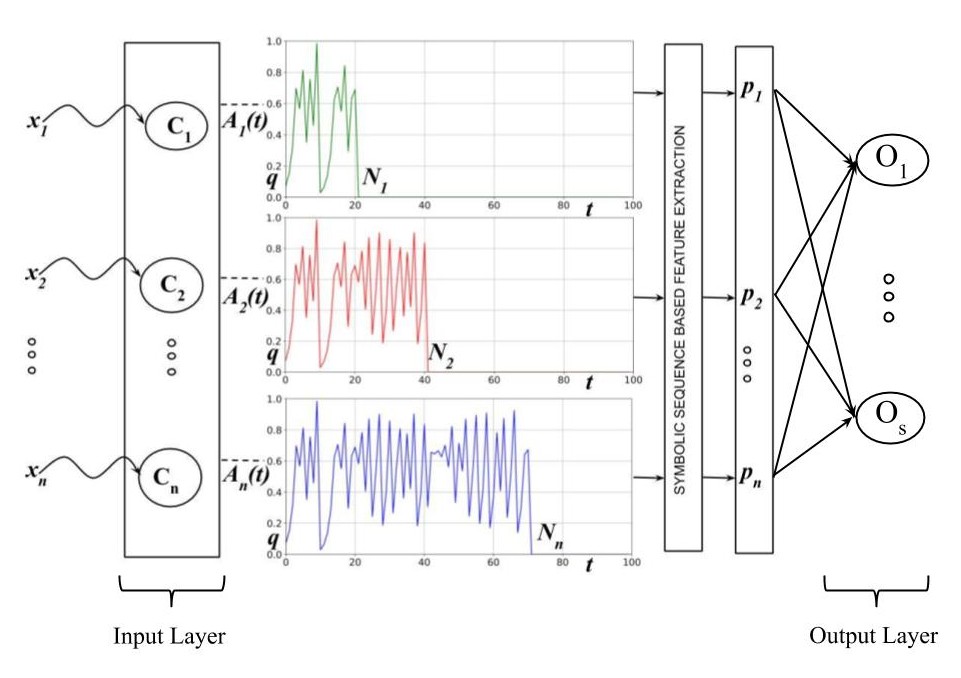}}
    
     \caption{ChaosNet: The proposed Chaotic GLS neural network architecture for classification tasks. $C_1, C_2,..., C_{n}$ are the 1D GLS chaotic neurons. The initial normalized neural activity of each neuron is $q$ units. The input or the normalized set of stimuli to the network is represented as $\{ x_i\}_{i=1}^{n}$. The chaotic firing of a GLS neuron $C_i$ halts when its chaotic activity value $A_{i}(t)$ starting from initial neural activity ($q$) reaches the $\epsilon$-neighbourhood of stimulus. This neuron has a {\it firing time} of $N_i$ $ms$. $A_{i}(t)$ contains 
     Topological transitivity symbolic sequence feature $p_i$. This feature is extracted from $A_{i}(t)$ of the $C_i$-th GLS-Neuron.}
    \label{model_archi}
    \end{figure}

We propose for the first time, a single layer chaos inspired neuronal architecture for solving classification problems (Figure~\ref{model_archi}). It consists of a single input and a single output layer. The input layer consists of $n$ GLS neurons $C_1, C_2, \ldots, C_n $ and extracts patterns from each sample of the input data instance. The nodes $O_1, O_2, \ldots, O_s $ in the output layer stores the representation vectors corresponding to each of the $s$ target classes ($s$-class classification problem). The entire input data is represented as a matrix (\emph{\textbf{X}}) of dimensions $m \times n$ where $m$ represents the number of data instances and $n$ represents the number of samples for each data instance. When the input data consists of $m$ images, each of dimensions $W \times Y$, we vectorize each image and the resulting matrix \emph{\textbf{X}} will be of dimensions $m \times n$ with $n=WY$. Each row of this matrix will be an instance of the vectorized image. The number of neurons in the input layer of \verb+ChaosNet+ is set equal to the number of samples of data instance.


The GLS neurons ($C_1, C_2, \ldots, C_n$) in the architecture in Figure~\ref{model_archi} have an initial neural activity of $q$ units. This is also considered as the initial value of the chaotic map. The GLS neurons starts firing chaotically when encountered by the stimulus. The stimulus is a real number which is normalized to lie between $0$ and $1$. The input vector of data samples (or data instance) represented as $x_1, x_2, \ldots, x_n$ in Figure~\ref{model_archi}
are the stimuli corresponding to $C_1, C_2, \ldots, C_n$ respectively. The chaotic neural activity values at time $t$ of the GLS neurons $C_1, C_2, \ldots, C_n$ are denoted by $A_{1}(t), A_{2}(t), \ldots, A_{n}(t)$ respectively, where
\begin{equation}
    A_{i}(t) = T(A_{i}(t-1)).
\end{equation}
The chaotic firing of each GLS neuron stops when their corresponding activity value $A_{1}(t), A_{2}(t), \ldots, A_{n}(t)$ starting from the initial neural activity ($q$) reaches the epsilon neighbourhood of $x_1, x_2, \ldots, x_n$. The time at which each neuron stops firing can thus be different. The halting of firing of each GLS neuron is guaranteed by {\it Topological Transitivity } (TT) property. The time taken ($N_{k}$ ms) for the firing of $k$-th GLS neuron to reach the epsilon neighbourhood of incoming stimulus is termed as {\it firing time}. The fraction of this firing time for which the activity of the GLS neuron is greater than the discrimination threshold ($b$) is defined as \emph{Topological Transitivity - Symbolic Sequence (TT-SS) feature}. The formal definition of {\it Topological Transitivity} is as follows:

\textbf{Definition 1:} {\it Topological Transitivity} property for a map $T: R \rightarrow R$ states that for all non-empty open set pairs $D$ and $E$ in $R$, there exists an element $d \in D$ and a non negative finite integer $n$ such that $T^{n}(d) \in E$.

For example, we consider a GLS-$1$D map ($T$) which is chaotic with $R: [0,1)$. Let $D= (q-\epsilon, q+\epsilon)$ and $E= (x_k-\epsilon, x_k+\epsilon)$ where $\epsilon > 0$. From the definition of {\it Topological Transitivity}, the existence of an integer $N_k \geq 0$ and a real number $d \in D$ such that $T^{N_k}(d) \in E$ is ensured. We consider $d = q$ (initial neural activity of the GLS-Neuron) and $x_k$ as the stimulus to the $k$-th GLS Neuron (after normalizing). Thus, $N_k \geq 0$ will always exist.  It is important to highlight that for certain values of $q$ there may be no such $N_k$, for eg., initial values that lead to periodic orbits. However, we can always find a value for $q$ for which $N_k$ exists. This is because, for a chaotic map, there are an infinite number of initial values that lead to non-periodic ergodic orbits.
%
%


\begin{figure}[htbp]
\centerline{ \includegraphics[width=0.6\textwidth]{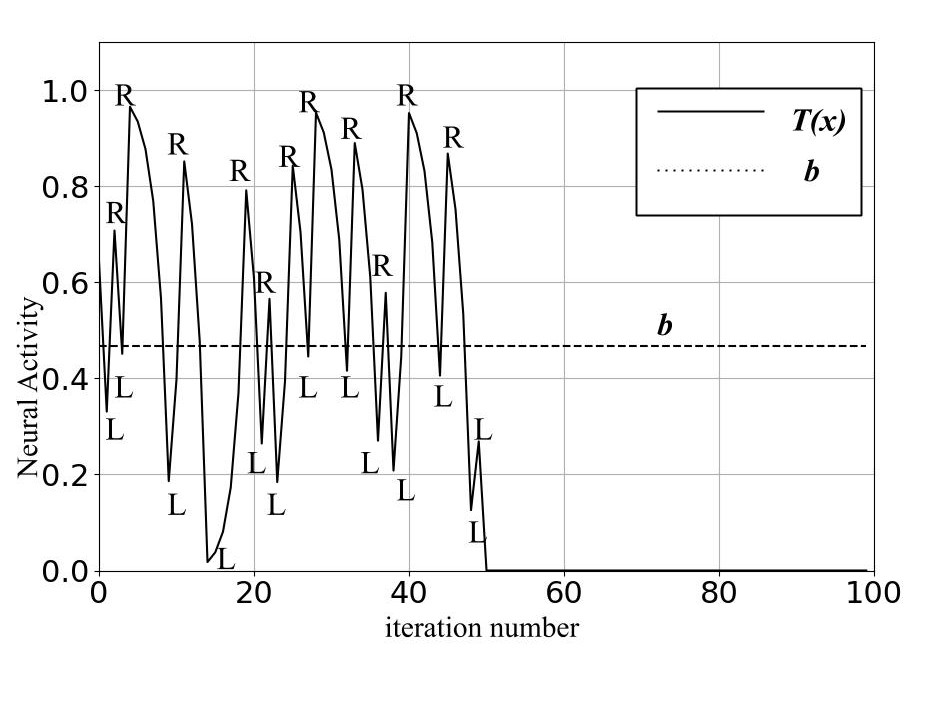}}

\caption{Illustration of feature extraction using Topological Transitivity - Symbolic Sequence method (Algorithm 2). The GLS-Neurons has a default initial neural activity of $q$ units. The $i$-th GLS-Neuron fires for $N_i$ iterations (chaotically) and stops by reaching the $\epsilon$-neighbourhood $I_{i}$ of the $i$-th input stimulus. For the duration of the GLS-Neuron being active (chaotic firing), the fraction of the time when the neural activity exceeds the internal discrimination threshold $b$ of the neuron (marked as R above) is the extracted TT-SS feature.}
\label{TT-SS-visual}
\end{figure}

\begin{figure*}
    \centering
    \includegraphics[scale=0.4]{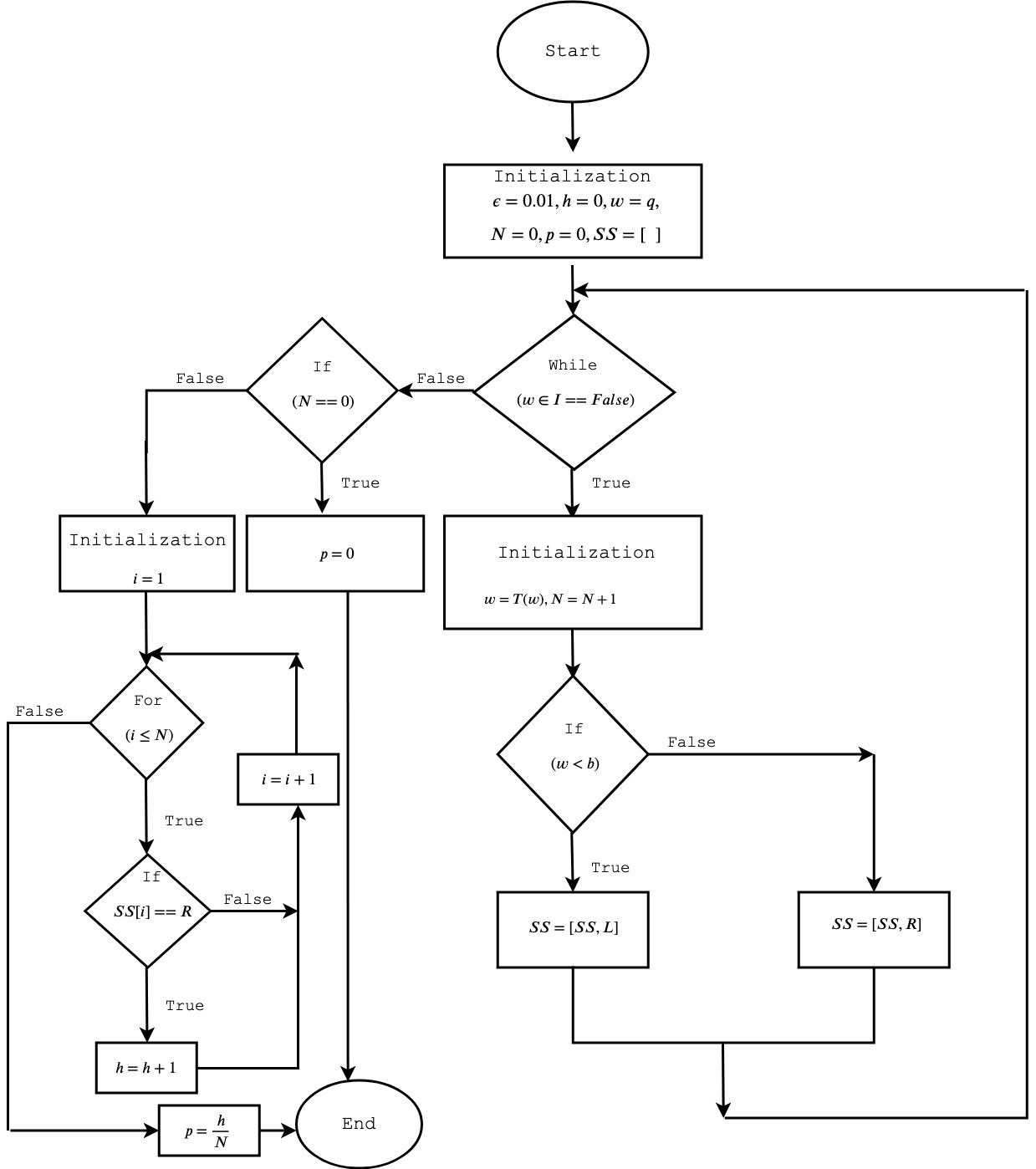}
    \caption{Flowchart for feature extraction step of 1-layer ChaosNet TT-SS algorithm.}
    \label{fig_ttss}
\end{figure*}

\begin{figure*}
    \centering
    \includegraphics[scale=0.4]{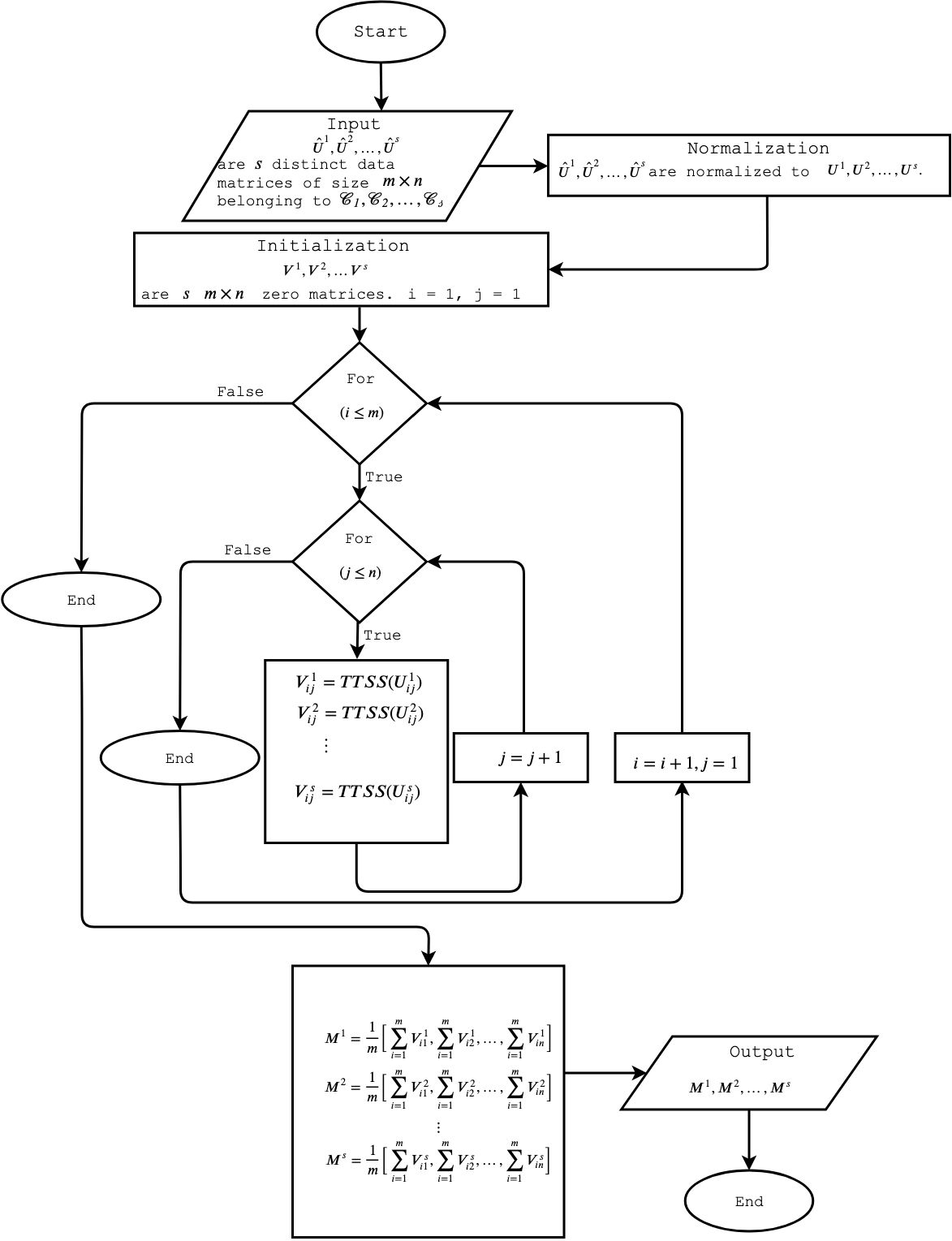}
    \caption{Flowchart for training step of 1-layer ChaosNet TT-SS algorithm.}
    \label{fig_training}
\end{figure*}

Single layer \verb+ChaosNet+ TT-SS algorithm consists of mainly three steps:

\begin{itemize}
\item \textbf{Feature extraction using TT-SS} -  TT-SS based feature extraction step is represented in the flowchart provided in Figure~\ref{fig_ttss}. Let the $\epsilon$-neighbourhood of $k$-th sample of $i$-th data instance $x_{k}^{i}$ be represented as $I_{k}^{i} = (x_{k}^{i}-\epsilon, x_{k}^{i}+\epsilon)$ where $\epsilon > 0$. Let the normalized stimulus to the $k$-th GLS-Neuron be $x_{k}^{i}$ and the corresponding neuronal {\it firing time}  be $N_{k}^{i}$. The firing trajectory of the GLS Neuron upon encountering a stimulus ($x^{i}_k$) is represented as $q \rightarrow T_{k}(q) \rightarrow T_{k}^{2}(q) \rightarrow T_{k}^{3}(q) \ldots \rightarrow T_{k}^{N_i}(q) $ where $T_{k}^{N_i}(q) \in I_{k}^{i}$. The trajectory is denoted as $\bm{A_{k}} = [q, T_{k}(q), \ldots, T_{k}^{N_i}(q)]$. The fraction of firing time for which the GLS-Neuron's chaotic trajectory activity value ($A_{k}(t)$) is greater than the internal discrimination threshold value ($b$) is defined as the Topological Transitivity - Symbolic Sequence (TT-SS) feature and denoted by $p^{i}_k$. 
    \begin{equation}
        p^{i}_k = \frac{h^{i}_k}{N^{i}_k},
    \end{equation}
    where $h^{i}_k$ represents the duration of firing for which the chaotic trajectory is above the discrimination threshold ($b$) for the $k$-th GLS Neuron (see  Figure~\ref{TT-SS-visual}). The TT-SS feature can be looked at as a spike-count rate based neural code~\cite{gerstner2014neuronal} for the GLS-Neuron which is active (or firing) for a total of $N_{k}^{i}$ time units, in which the spiking activity (greater than threshold activity) is limited to $h^{i}_k$ time units.
     \item \textbf{TT-SS Training} -  TT-SS based training step is represented in the flowchart provided in Figure~\ref{fig_training}. Let us assume an $s$ class classification problem where classes are represented by $\mathscr{C}_1$, $\mathscr{C}_2$, \ldots, $\mathscr{C}_s$ and the corresponding true-labels be denoted as $1, 2, \ldots, s$ respectively. Let the normalized distinct matrices be $U^1, U^2, \ldots,  U^s$ of size $m \times n$. The matrices $U^1, U^2, \ldots,  U^s$ denotes the data belonging to  $\mathscr{C}_1$, $\mathscr{C}_2$, \ldots, $\mathscr{C}_s$ respectively.
     Training involves extracting features from  $U^1, U^2, \ldots, U^s$ using TT-SS method so as to  yield $V^{1}, V^{2}, \ldots, V^{s}$. Feature extraction using TT-SS algorithm is applied on each stimulus, hence the size of $V^{1}, V^{2}, \ldots, V^{s}$ will be same as $U^{1}, U^{2}, \ldots, U^{s}$. Once the TT-SS based feature extraction is found for the data belonging to the $s$ distinct classes, the average across row is computed next:
\begin{eqnarray*}
    M^{1} &=& \frac{1}{m} \Big[ \sum_{i = 1}^{m} V^{1}_{i1}, \sum_{i = 1}^{m} V^{1}_{i2}, \ldots, \sum_{i = 1}^{m} V^{1}_{in} \Big], \\
M^{2} &=& \frac{1}{m} \Big[ \sum_{i = 1}^{m} V^{2}_{i1}, \sum_{i = 1}^{m} V^{2}_{i2}, \ldots, \sum_{i = 1}^{m} V^{2}_{in} \Big],\\
&\vdots&\\
M^{s} &=& \frac{1}{m} \Big[ \sum_{i = 1}^{m} V^{s}_{i1}, \sum_{i = 1}^{m} V^{s}_{i2}, \ldots, \sum_{i = 1}^{m} V^{s}_{in} \Big].
\end{eqnarray*}
$M^1, M^2, \ldots, M^s$ are $s$ row vectors and are termed as {\it mean representation vectors} corresponding to the $s$ classes.
$M^{k}$ is a vector where the {\it average internal representation} of all the stimuli corresponding to $k$-th class is encoded. As more and more input data are received the mean representation vectors get updated. 


\end{itemize}

\begin{figure*}
    \centering
    \includegraphics[scale=0.4]{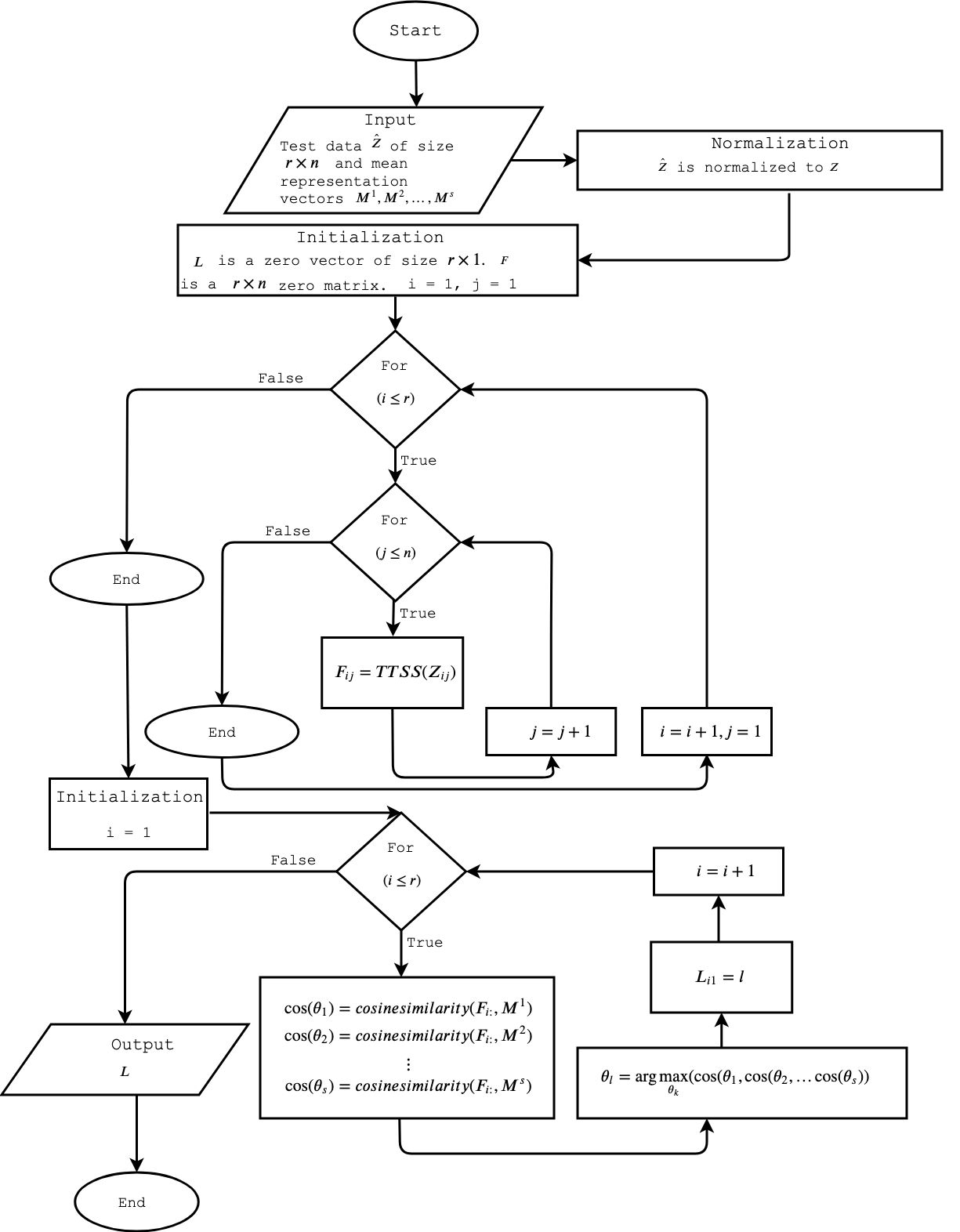}
    \caption{Flowchart for testing step of 1-layer ChaosNet TT-SS algorithm.}
    \label{fig_testing}
\end{figure*}

\begin{itemize}

\item \textbf{TT-SS Testing} - The computational steps involved in testing are in the flowchart provided in Figure~\ref{fig_testing}. Let $Z$ denote the normalized test data matrix of size  $r \times n$. The $i$-th row of $Z$ represents $i$-th test data instance denoted as $z^i = [z^{i}_{1}, z^{i}_{2}, z^{i}_{3}, \ldots, z^{i}_{n}]$. The TT-SS based feature extraction step is applied to each row (each test data instance $(z^i)$ where $i = 1, 2, 3, \ldots, r$). Let the feature extracted data of test samples using TT-SS algorithm be denoted as $F$ where $f^{i} =[f^{i}_1, f^{i}_2, \ldots, f^{i}_n]$ is the $i$-th row of $F$.
Feature extraction is followed by the computation of cosine similarity of $f^{i}$ independently with each of the $M^1, M^2,..., M^s$ ({\it mean representation} vectors) respectively:
  %
%
\begin{eqnarray*}
\cos(\theta_{1}) &=& \frac{f^{i} \cdot M^{1}}{\left \|f^{i} \right \|_2 \left \|M^{1} \right \|_2}, \\
\cos(\theta_{2}) &=& \frac{f^{i} \cdot M^{2}}{\left \|f^{i} \right \|_2 \left \|M^{2} \right \|_2},\\
&\vdots&\\
\cos(\theta_{s}) &=& \frac{f^{i} \cdot M^{s}}{\left \|f^{i} \right \|_2 \left \|M^{s} \right \|_2},
\end{eqnarray*}

$f^i \cdot M^k$ represents the scalar product between vectors $f^i$ and $M^k$, $\left \|v \right \|_2$ denotes the $l_2$ norm of row-vector $v$. The above will give $s$ scalar values which are the cosine similarity values between $\{ M^k \}_{k=1}^{s}$ and $f^i$. Out of these $s$ scalar values, the index ($l$) corresponding to the maximum cosine similarity ($\cos(\theta_{l})$) is considered as the label for $f^{i}$:
\begin{equation*}
      \theta_l  = \arg\max_{\theta_i}  (\cos(\theta_{1}), \cos(\theta_{2}), \ldots, \cos(\theta_{s})).
\end{equation*}
If there is more than one index with maximum cosine similarity, we take the smallest such index. The above procedure is continued until a unique label is assigned to each test data instance.
\end{itemize}
\subsection{Parameters vs Hyperparameters}
Distinguishing model parameters and model hyperparameters plays a crucial role in machine learning tasks. The model parameters and hyperparameters of the proposed method are as follows:

{\it Model parameter}: This is an internal parameter which is estimated from the data. These internal parameters are what the model learns while training. In the case of single layer \verb+ChaosNet+ TT-SS method, the { \it mean representation vectors } $M^{1}, M^{2}, \ldots, M^{s}$ are the model parameters which are learned while training. The model parameters for deep learning (DL) are the weights and biases learnt during training the neural network. In Support Vector Machines (SVM), the support vectors are the parameters. In all these cases, the parameters are learned while training.

{\it Model hyperparameters}: These are configurations which are external to the model and typically not estimated from the data. The hyperparameters are tuned for a given classification or predictive modelling task in order to estimate the best model parameters. The hyperparameters of a model are often arrived by heuristics and hence are different for different tasks. In the case of single layer \verb+ChaosNet+ TT-SS method, the hyperparameters are the {\it initial neural activity} ($q$), {\it internal discrimination threshold} ($b$), $\epsilon$ used in defining the neighbourhood interval of $x^{i}_{k}$  ($I_{k}^{i} = (x_{k}^{i} - \epsilon, x_{k}^{i} + \epsilon)$) and the {\it chaotic map} chosen. In DL, the hyperparameters are the {\it number of hidden layers},  {\it number of neurons in the hidden layer}, {\it learning rate} and the {\it activation function} chosen. In the case of K-Nearest Neighbours (KNN) classification algorithm, the {\it number of nearest neighbours} ($k$) is a hyperparameter. In SVM, the {\it choice of kernel} is a hyperparameter. In Decision Tree, {\it depth of the tree} and the {\it least number of samples required to split the internal node}  are the hyperparameters. In all these cases the hyperparameters need to be fixed by cross-validation.

\subsection{Example to illustrate single layer ChaosNet TT-SS method}
We consider a binary classification problem. The two classes are denoted as $\mathscr{C}_1$ and $\mathscr{C}_2$ with class labels $1$ and $2$ respectively. The input dataset is a $4 \times 4$ matrix $\hat{X} = \big[\frac{\hat{X}^1}{\hat{X}^2}\big]$. The first two rows of $\hat{X}$ are denoted as $\hat{X}^1$ and the remaining two rows of $\hat{X}$ are denoted as $\hat{X}^2$. $\hat{X}^1$ and $\hat{X}^2$ represent data instances belonging to $\mathscr{C}_1$ and $\mathscr{C}_2$ respectively. Since each row of $\hat{X}$ has $4$ samples, the input layer of \verb+ChaosNet+ in this case has $4$ GLS neurons $\{C_1, C_2, C_3, C_4\}$ in the input layer. We set the hyperparameters $q = 0.23$, $b = 0.56$, $\epsilon = 0.01$ and chaotic map as $T_{Skew-Tent}$. 
The steps of the $1$-layer \verb+ChaosNet+ TT-SS algorithm are:
 \begin{enumerate}
    \item Step 1: Normalization of data: $\hat{X}$ is normalized\footnote{For a non-constant matrix $X$, normalization is achieved by performing $\frac{X-\min(X)}{\max(X)-\min(X)}$. A constant matrix $X$ is normalized to all ones.} to $X$.
    
    \item Step 2: Training - TT-SS based feature extraction: From the normalized matrix $X$, we sequentially pass one row of $X$ at a time to the input layer of \verb+ChaosNet+ and extract the TT-SS feature. As an example, let $x^1 = [0.2, 0.5, 0.1, 0.9]$ and $x^2 = [0.23, 0.49, 0.15, 0.8]$ be the first two rows of $X$ (pertaining to $\mathscr{C}_1$) which are passed to the input layer sequentially. Each GLS neuron in the input layer $\{C_1, C_2, C_3, C_4\}$ starts firing chaotically until it reaches the $\epsilon$-neighbourhood of $x^1$ which are $(0.2-\epsilon, 0.2 + \epsilon)$, $(0.5-\epsilon, 0.5 + \epsilon)$ , $(0.1-\epsilon, 0.1 + \epsilon)$ and $(0.9-\epsilon, 0.9 + \epsilon)$. Let the neural firing of $\{C_1, C_2, C_3, C_4\}$ be denoted as $\bm{A_{1}}$, $\bm{A_{2}}$, $\bm{A_{3}}$, $\bm{A_{4}}$. The fraction of firing time for which $A_{i}(t) > b$ (internal discrimination threshold) is determined for the $4$ GLS neurons to yield the TT-SS feature vector $v^1 = [0.41, 0.35, 0.38, 0.5]$. Similarly, the TT-SS feature vector for $x^{2}$ is $v^2 = [0, 0.35, 0.35, 0.41]$. The mean representation vector of $\mathscr{C}_1$ is calculated as $M^{1} = [\frac{(0.41 + 0)}{2}, \frac{(0.35 + 0.35)}{2}, \frac{(0.38 + 0.35)}{2}, \frac{(0.5 + 0.41)}{2} ] = [0.205, 0.35, 0.365, 0.455]$. In a similar fashion, the mean representation vector of $\mathscr{C}_2$ is computed. Thus, at the end of training, the mean representation vectors $M^{1}$ and $M^{2}$ of $\mathscr{C}_1$ and $\mathscr{C}_2$ are stored in nodes $O_1$ and $O_2$ of the output layer respectively.
    
    \item Step 3: Testing - as an example, let $z = [z_1, z_2, z_3, z_4] $ be a test sample (after normalization) that needs to be classified. Similar to the training step, $z$ is first fed to the input layer of \verb+ChaosNet+ having 4 GLS neurons and its activity is recorded. Subsequently, TT-SS feature vector is extracted from the neural activity which is denoted as $f = [f_1, f_2, f_3, f_4]$. We compare $f$ individually with the two internal representation vectors $M^{1}$ and $M^{2}$ by using cosine similarity metric. The test sample $z$ is classified to that class for which the cosine similarity is maximum.     
\end{enumerate}

A single-layer \verb+ChaosNet+ with finite number of GLS neurons has the ability to approximate any arbitrary  real-valued, bounded discrete time function (with finite support) as we shall show in the next subsection.

\subsection{\label{sec:UAT} Universal Approximation Theorem (UAT)} 
Cybenko (in 1989) proved one of the earliest versions of the {\it Universal Approximation Theorem} (UAT). {\it UAT} states that continuous functions on compact subsets of $\mathbb{R}^n$ can be approximated by an ANN with $1$- hidden layer having finite number of neurons with sigmoidal activation functions~\cite{cybenko1989approximation}. Thus, simple neural networks with appropriately chosen parameters can approximate continuous functions of a wide variety. We have recently proven a version of this theorem for the GLS-neuron~\cite{kathpalia2019novel} which we reproduce below (with minor modifications).\\

\noindent \par \textbf{Universal Approximation Theorem (UAT) for GLS-Neuron:~}

Let us consider a real valued bounded discrete time function $g(n)$ having a support length $LEN$. {\it UAT} guarantees the existence of a finite set of GLS neurons denoted as $W_1$, $W_2$, $\ldots$, $W_c$, such that, for these $c$ neurons, the GLS-encoded output values $x_1,p_1$, $x_2,p_2$, $\ldots$, $x_c, p_c$ can approximate $g(n)$. Here $x_i$s are the initial values and $p_i$s are the skew parameters of the GLS maps. In other words, the {\it UAT} for GLS neurons guarantees the existence of the function 
$G(\cdot)$ satisfying the following:
\begin{equation}
|G(x_1,x_2,\ldots,x_c,p_1, p_2, \ldots, p_c) - g(n)| \leq \epsilon,
\label{eq_UAT}
\end{equation}
where $\epsilon>0$ is arbitrarily small. 
\\

\noindent {\textit{Proof:}}~By construction: for a given $\epsilon > 0$, the range of the function $g$ is uniformly quantized in such a way that the quantized and original function differ by an error $\leq \epsilon$. The boundedness of $g(n)$ ensures that the above is always true, because we can find integers $a_1$, $a_2$, $\ldots$, $a_{LEN}$ corresponding to the time indices $1, 2, \ldots, LEN-1, LEN$ which satisfy the following inequality after proper global scaling of $g$: 
\begin{equation}
\frac{1}{S_{\epsilon}}  \sum_{i=1}^{i=L} \vert (g^{*}(i) - a_i) \vert \leq \epsilon,
\end{equation}
where $\epsilon > 0$, $a_i$-s are all integers and $g^{*}(\cdot) = S_{\epsilon} g(\cdot)$, where $S_{\epsilon}$ (a finite real number) denotes the proper global scaling constant which in turn depends on $\epsilon$. Let us consider only the significant number of bitplanes of $a_i$-s - denote it as  $c$ (the value of $c$ is the least power of 2 which is just greater than the maximum of $\{ a_i \}$).

Let the discrete-time quantized integer-valued signal that approximates $g(n)$ be called as $g_{quant}[n] = a_1, a_2, \ldots, a_{LEN}$. The $c$ bitplanes of each value of the function $g_{quant}[n]$ are extracted next to yield $c$ bitstreams $g_c[n], g_{c-1}[n],$ $\ldots, g_1[n]$, where the Most Significant Bit (MSB) and Least Significant Bit (LSB) are denoted by $c$ and $1$ respectively. The $i$-th stream of bits is a $LEN$ length binary list. The back-iteration on an appropriate GLS can encode the binary list {\it losslessly} to the initial value $x_i$ (the probability of 0 in the bitstream is represented by the parameter $p_i$ which is also the skew parameter of the map). The above procedure is followed for the  $c$ bitstreams to yield $x_1,p_1$, $x_2,p_2$, $\ldots$, $x_c, p_c$ GLS encoded neurons. The perfect lossless compression property of GLS enables the recovering of original quantized  bitstreams using GLS decoding~\cite{nagaraj2009arithmetic}. Each step in our construction uses procedures that are deterministic (and which always halt). This means the compositions of several non-linear maps (quantization, scaling, bitplane extraction, GLS encoding are all non-linear) can be used to construct the desired function $G$. Thus, there exists a function $G(x_1,x_2,\ldots,x_c,p_1, p_2, \ldots, p_c)$ that satisfies inequality (Eq.~\ref{eq_UAT}). \hfill $\square$

 The function $G(x_1,x_2,\ldots,x_c,p_1, p_2, \ldots, p_c)$ is not unique since it depends on how the $a_i$-s are chosen and finding an explicit analytical expression is not easy (but we know it exists). The above argument can be extended for continuous-valued functions (by sampling) as well as functions in higher dimensions.

\section{\label{sec:Experiments} Experiments, Results and Discussion}
Learning from insufficient training samples is very challenging for ANN/ML/DL algorithms since they are typically dependent on learning from vast amount of data. The efficacy of our proposed \verb+ChaosNet+ TT-SS method based classification is evaluated on MNIST, KDDCup'99, Iris and Exoplanet data in the low training sample regime. The description of the datasets used in the analysis of the proposed method are given here.
\subsection{Datasets}
\subsubsection{MNIST}
MNIST~\cite{lecun-mnisthandwrittendigit-2010} is a commonly used hand written digits ($0$ to $9$) image dataset in the ML/DL community. These images have a dimension of 28 pixels $\times$ 28 pixels and are stored digitally as $8$-bit grayscale images. The training set of MNIST database consists of $60,000$ images whereas the test set consists of $10,000$ images. MNIST is a multi-class classification ($10$ classes) problem and the goal is to classify the images  into their correct respective classes.
For our analysis, we have independently trained with randomly chosen $1, 2, 3, \dots, 21$ data samples per class. For each such trial of training, the algorithm is tested with ($10,000$) unseen test images.


\subsubsection{KDDCup'99}
KDDCup'99~\cite{cup1999data} is a benchmark dataset used in the evaluation of intrusion detection systems (IDS). The creation of this dataset is based on the acquired data in IDS DARPA'98 evaluation program~\cite{lippmann2000evaluating}. 
There are roughly 4,900,000 single connection vectors in the KDDCup'99 training data. The number of features in each connection vector is 41. Each data sample is labeled as either {\it normal} or a {\it specific attack type}. We considered 10\% of data samples from the entire KDDCup'99 data. In this 10\% KDDCup'99 data that we considered, there are normal and 21 different attack categories. Out of these, we took the following $9$ classes for our analysis: {\it back}, {\it ipsweep}, {\it neptune}, {\it normal}, {\it portsweep}, {\it satan}, {\it smurf}, {\it teardrop}, and {\it warezmaster}. Training was done independently with $1, 2, 3, \dots, 7$ data samples per class. The data samples were chosen randomly for training from the existing training set. For each trial of training, the algorithm is tested with unseen data.

\subsubsection{Iris}
Iris data\footnote{\url{http://archive.ics.uci.edu/ml/datasets/iris}}~\cite{blake1998uci} consists of attributes from 3 types of Iris plants. These plants are as follows: {\it Setosa}, {\it Versicolour} and {\it Virginica}. The number of attributes used in this dataset are 4. The attributes are {\it sepal length}, {\it sepal width}, {\it petal length} and {\it petal width} (all measured in centimeters). This is a $3$ class classification problem with $50$-data samples per category. For our analysis, we have independently trained with randomly chosen $1, 2, 3, \dots, 6, 7$ data samples per class. For each trial of training, the algorithm is tested with ($120$) unseen test data.



\subsubsection{Exoplanet}
PHL-EC dataset\footnote{The habitable Exoplanet Catalog: \url{http://phl.upr.edu/hec}} (combined with stellar data from the Hipparcos catalog~\cite{Mendez_hipp} ) has 68 attributes (of which 55 are continuous valued and 13 are categorical) and more than $3800$ confirmed exoplanets (at the time of writing this paper). Important attributes such as surface temperature, atmospheric type, radius, mass, flux, earth's similarity index,  escape velocity, orbital velocity etc. are included in the catalog (with both observed and estimated attributes). From an analysis point-of-view, this presents interesting challenges~\cite{SBAF, ASCOM18}. The dataset consists of  six classes, of which three were used in our analysis as they are sufficiently large in size. The other three classes are dropped (while training) since they had very low number of samples. The classes considered are  {\it mesoplanet}, {\it psychroplanet} and {\it non-habitable}. Based on their thermal properties,  these three classes or types of planets are described as follows:
%
%
%
%
%
\begin{enumerate}
\item \textbf{Mesoplanets}: Also known as M-planets, these have mean global surface temperature in the range 0$^\circ$C to 50$^\circ$C which is a necessary condition for survival of complex terrestrial life. The planetary bodies with sizes smaller than Mercury and larger than Ceres fall in this category, and these are generally referred to as Earth-like planets.
%

\item \textbf{Psychroplanets}:  Planets in this category have mean global surface temperature in the range -50$^\circ$C to 0$^\circ$C. This is much colder than optimal for terrestrial life to sustain. 
%

\item \textbf{Non-Habitable}: Those planets which do not belong to either of the above two categories fall in this category. These planets do not have necessary thermal properties for sustaining life.
%
\end{enumerate}
The three remaining classes in the data are -- thermoplanet, hyperthermoplanet and hypopsychroplanet. However, owing to highly limited number of samples in each of these classes, we ignore these classes. While running the classification methods, we consider multiple attributes of the parent sun of the exoplanets that include mass, radius, effective temperature, luminosity, and the limits of the habitable zone.
The first step consists of pre-processing data from PHL-EC. An important challenge in the dataset is that a total of about $1\%$ of the data is missing (with a majority being of the attribute P. Max Mass) and in order to overcome this, we used a simple method of removing instances with missing data after extracting the appropriate attributes for each experiment, as most of the missing data is from the non-habitable class. We considered another data set where a subset of attributes (restricted attributes) consisting of Planet Minimum Mass, Mass, Radius, SFlux Minimum, SFlux Mean, SFlux Maximum are used as input. This subset of attributes do not consider surface temperature and any attribute related to surface temperature at all, making the decision boundaries more complicated to decipher. Following this, the ML approaches were used on these preprocessed datasets. The online data source for the current work is available at \url{http://phl.upr.edu/projects/habitable-exoplanets-catalog/data/database}.


\subsection{Results and Discussion}
We compare the proposed $1$-layer \verb+ChaosNet+ TT-SS method with the following ML algorithms:  Decision Tree~\cite{quinlan1986decision_tree}, K-Nearest Neighbour (KNN~\cite{cover1967nearest}), Support Vector Machine (SVM~\cite{hearst1998support}) and Deep Learning algorithm (DL, 2 layers~\cite{lecun2015deep}). The parameters used in ML algorithms are provided in \textit{Appendix}. We have used {\it Scikit-learn}~\cite{scikit-learn} and {\it Keras}~\cite{chollet2015keras} package for the implementation of ML algorithms and $2$-layer neural network respectively.
\begin{table}[h!]
\centering
\caption{Hyperparameter settings -- initial neural activity, discrimination threshold value, type of GLS-Neuron and value of $\epsilon$ used in the study.}
\vspace{0.1in}
\begin{tabular}{lccccr}
\hline
\hline
Dataset & Initial Neural & Discrimination & GLS-Neuron & $\epsilon$ \\ 
 & Activity ($q$) & Threshold ($b$) & type & \\
\hline
MNIST & $0.3360$ & $0.3310000$ & $T_{Skew-Binary}$ & $0.01$ \\ 
KDDCup'99 & $0.6000$ & $0.3350000$ & $T_{Skew-Tent}$ & $0.01$ \\ 
Iris & $0.6000$ & $0.9867556$ & $T_{Skew-Binary}$ & $0.01$ \\ 
Exoplanet & $0.2624$\tablefootnote{Actual value was $0.26242424242424245$.}  & $0.1490000$ & $T_{Skew-Tent}$ & $0.01$ \\ 
Exoplanet\tablefootnote{Exoplanet with No Surface Temperature} & $0.2624$\tablefootnote{Actual value was $0.26242424242424245$.} & $0.1490000$ & $T_{Skew-Tent}$ & $0.01$ \\ 
Exoplanet\tablefootnote{Exoplanet with Restricted attributes} & $0.9500$\tablefootnote{Actual value was $0.9500000000000006$.}& $0.4760000$ & $T_{Skew-Tent}$ & $0.001$ \\ 
\hline
\hline
\end{tabular}
\label{table_parameters_ttss}
\end{table}
%
\begin{figure}[h!]
    \centerline{ \includegraphics[width=0.45\textwidth]{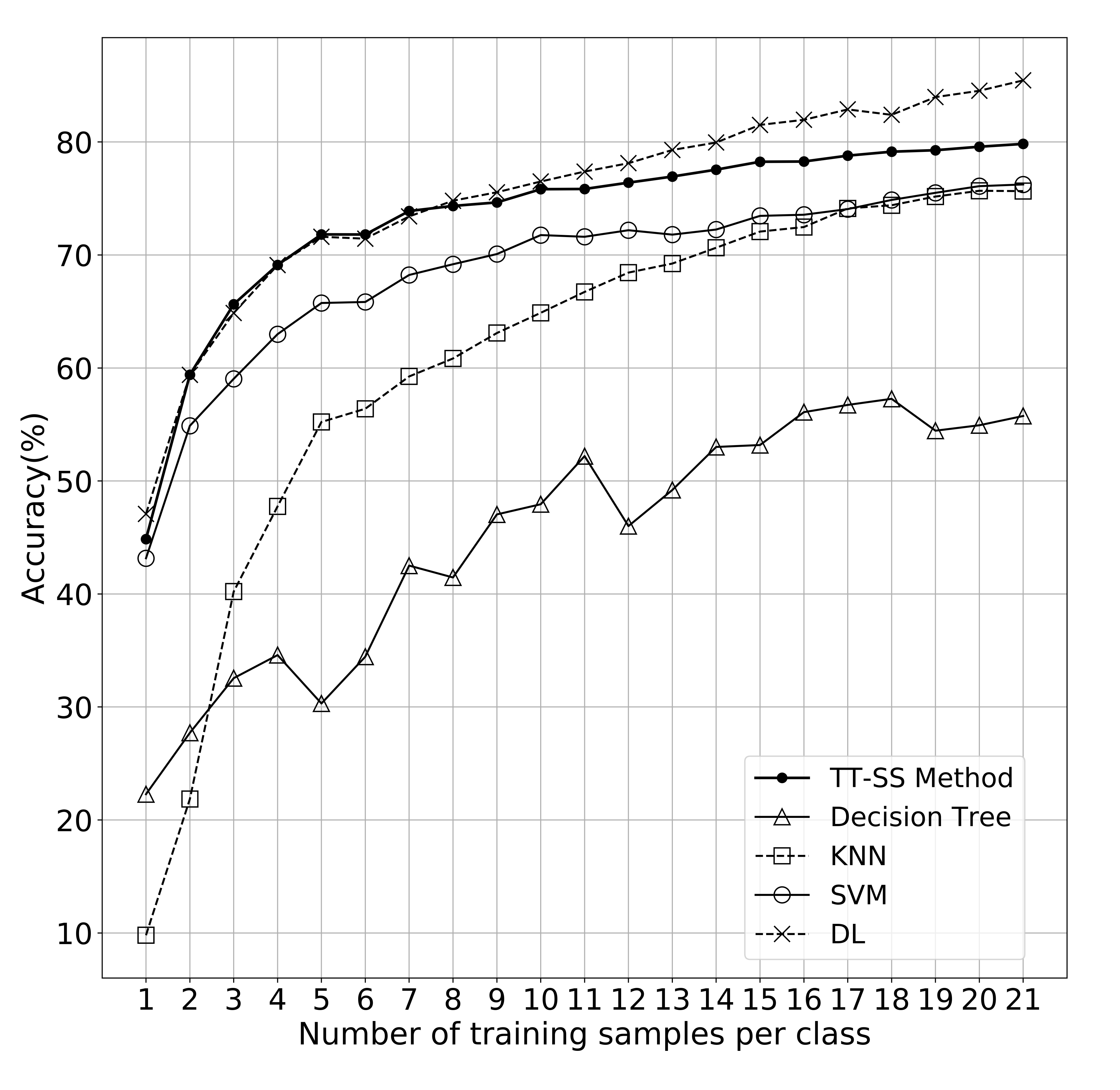}}
    
    \caption{ Performance comparison of ChaosNet TT-SS algorithm with DL ($2$-layers), Decision Tree, SVM and KNN for MNIST dataset in low training sample regime. }
    \label{ttssvsML_mnist}
    \end{figure}
\begin{figure}[h!]
    \centerline{ \includegraphics[width=0.45\textwidth]{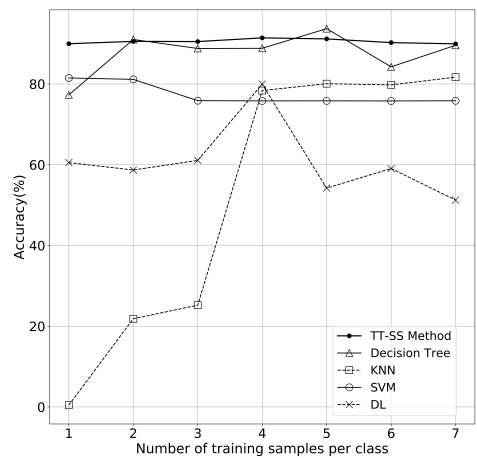}}
    \caption{ Performance comparison of ChaosNet TT-SS algorithm with DL ($2$-layers), Decision Tree, SVM and KNN for KDDCup'99 dataset in the low training sample regime.}
    
    \label{ttssvsML_kddcup}
    \end{figure}

\begin{figure}[h!]
    \centerline{ \includegraphics[width=0.45\textwidth]{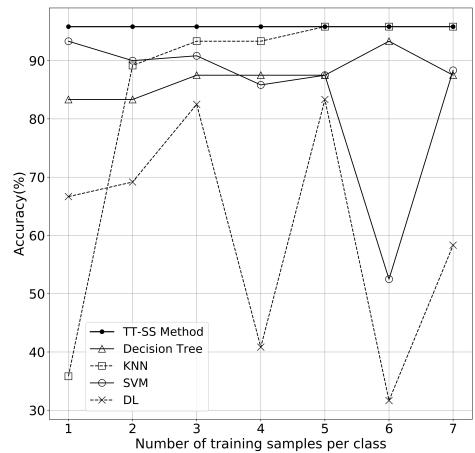}}
    
    \caption{ Performance comparison of ChaosNet TT-SS algorithm with DL ($2$-layers), Decision Tree, SVM and KNN for Iris dataset in the low training sample regime.}
    \label{ttssvsML_iris}
    \end{figure}

\begin{figure}[h!]
    \centerline{ \includegraphics[width=0.45\textwidth]{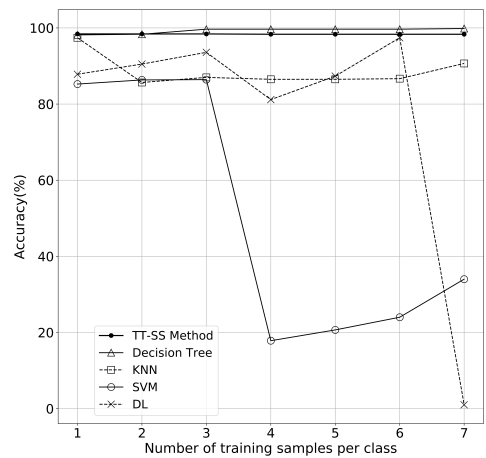}}
    
    
    \caption{ Performance comparison of ChaosNet TT-SS algorithm with DL ($2$-layers), Decision Tree, SVM and KNN for Exoplanet dataset in the low training sample regime.}
    \label{ttssvsML_exoplanet}
    \end{figure}

\begin{figure}[h!]
    \centerline{ \includegraphics[width=0.45\textwidth]{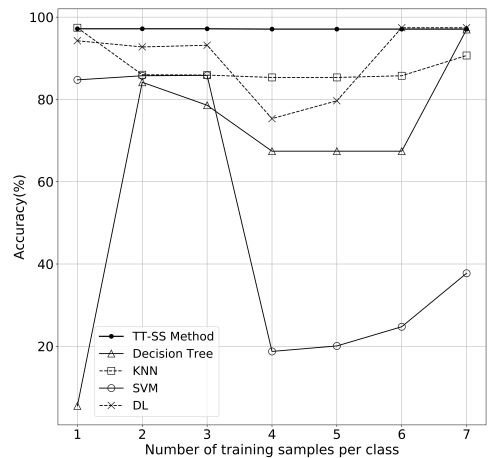}}
    
      \caption{ Performance comparison of ChaosNet TT-SS algorithm with DL ($2$-layers), Decision Tree, SVM and KNN for Exoplanet dataset with no surface temperature in the low training sample regime.}
    \label{ttssvsML_exoplanet_no_stemp}
    \end{figure}
    
\begin{figure}[h!]
    \centerline{ \includegraphics[width=0.45\textwidth]{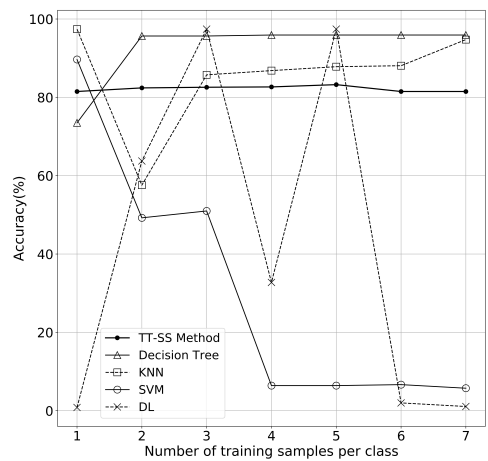}}
    
      \caption{ Performance comparison of ChaosNet TT-SS algorithm with DL ($2$-layers), Decision Tree, SVM and KNN for Exoplanet dataset with 6 restricted attributes in the low training sample regime.}
    \label{ttssvsML_exoplanet_restricted}
    \end{figure}    

%

%
%
{\it Performance of \verb+ChaosNet+ TT-SS Method on MNIST data}:
Figure~\ref{ttssvsML_mnist} shows the comparative performance of \verb+ChaosNet+ TT-SS method with SVM, Decision Tree, KNN and DL (2-layer) on MNIST dataset. In the case of MNIST data \verb+ChaosNet+ TT-SS method outperforms classsical ML techniques like SVM, Decision Tree and KNN. The \verb+ChaosNet+ TT-SS method gave slightly higher performance than DL upto training with 8 samples per class. As the number of training samples increased (beyond 8), DL outperforms \verb+ChaosNet+ TT-SS method.

{\it Performance of \verb+ChaosNet+ TT-SS Method on KDDCup'99 data}: 
Figure~\ref{ttssvsML_kddcup} shows the comparative performance of \verb+ChaosNet+ TT-SS method with SVM, Decision Tree, KNN and DL (2-layer) on KDDCup'99 dataset. In the low training sample regime for KDDCup'99 data, \verb+ChaosNet+ TT-SS method outperforms the classical ML and DL algorithms except for training with $5$ samples/class, where Decision Tree outperforms \verb+ChaosNet+ TT-SS method.

{\it Performance of \verb+ChaosNet+ TT-SS Method on Iris data}: 
Figure~\ref{ttssvsML_iris} shows the comparative performance of \verb+ChaosNet+ TT-SS method with SVM, Decision Tree, KNN and DL (2-layer) on Iris dataset. In the case of Iris data, \verb+ChaosNet+ TT-SS method gives consistently the best results when trained with $1, 2, \ldots, 7$ samples per class.

{\it Performance of \verb+ChaosNet+ TT-SS Method on Exoplanet data}: Figure~\ref{ttssvsML_exoplanet} shows the comparative performance of \verb+ChaosNet+ TT-SS method with SVM, Decision Tree, KNN and DL (2-layer) on Exoplanet dataset. We consider $3$ classes from the exoplanet data, namely Non- Habitable, Meso Planet and Psychroplanet. There are a total of $45$ attributes to explore the classification of $3$ types of exoplanet. We have considered another scenario where surface temperature (Figure~\ref{ttssvsML_exoplanet_no_stemp}) is removed from the set of attributes. This makes the classification problem harder as the decision boundaries between classes become fuzzy in the absence of surface temperature. Additionally, restricted set of attributes (Figure~\ref{ttssvsML_exoplanet_restricted}) is considered where the direct/indirect influence of surface temperature is mitigated by removing all related attributes from the original full set of attributes. This makes habitability classification an incredibly complex task. Even though the literature is replete with possibilities of using both supervised and unsupervised learning methods, the soft margin between classes, namely psychroplanet and mesoplanet makes the task of discrimination incredibly difficult. This has perhaps resulted in very few published work on automated habitability classification. A sequence of recent explorations by Saha et. al. (2018) expanding previous work by Bora et. al \cite{CDHS} on using Machine Learning algorithm to construct and test planetary habitability functions with exoplanet data raises important questions.

In our study, independent trials of training is done with very less samples: $1, 2, \ldots, 7$ randomly chosen data samples per class. The algorithm is then tested on unseen data for each of the independent trials. This accounts for the efficacy of the algorithm in detecting variances in new data. A consistent performance of \verb+ChaosNet+ TT-SS method is observed in the low training sample regime. \verb+ChaosNet+ TT-SS method gives the second highest performance in terms of accuracy when compared to SVM, KNN and DL. The highest performance is given by Decision Tree. Despite the sample bias due to the non-habitable class, we were able to achieve remarkable accuracies with the proposed algorithms without having to resort to under sampling and synthetic data augmentation. Additionally, the performance of \verb+ChaosNet+ TT-SS is consistent compared to other methods used in the analysis. 

\subsection{Single layer ChaosNet TT-SS algorithm in the presence of Noise}
One of the key ways to identify the robustness of a ML algorithm is by testing its efficiency in classification or prediction in the presence of noise. Robustness needs to be tested under the following scenarios:  noisy test data, training data attributes affected by noise, inaccurate training data labels due to the influence of noise and noise affected model parameters\footnote{Hyperparameters are rarely subjected to noise and hence we ignore this scenario. It is always possible to protect the hyperparameters by using strong error correction codes.}. Amongst these, {\it noise affected model parameters} is the most challenging since it can significantly impact performance of the algorithm. We consider a scenario where the parameters learned by the model while training are passed through a noisy channel. As an example, we compare the performance of the single layer \verb+ChaosNet+ TT-SS algorithm and $2$-layer neural network (DL architecture) for Iris data. The parameters for the both algorithms are modified by Additive White Gaussian Noise (AWGN) with zero mean and increasing variance. The Iris dataset consists of $3$ classes with $4$ attributes per data instance. We considered the specific case of training with only $7$ samples per class. 

{\it Parameters settings for the single layer TT-SS algorithm implemented on \verb+ChaosNet+ for Iris data:} Corresponding to the 3 classes for the Iris data, the output layer of \verb+ChaosNet+ consists of $3$-nodes $O_1$, $O_2$, and $O_3$ which store the mean representation vectors. Since each representation vector contains 4 components (corresponding to the 4 input attributes), the total number of learnable parameters are $12$. These parameters are passed through a channel corrupted by AWGN with zero mean and increasing  standard deviation (from $0.0001$ to $0.0456$). This results in a variation of the Signal-to-Noise Ratio (SNR) from $-6.98$ $dB$ to $45.36$ $dB$. 

{\it Parameter settings for the $2$-layer neural network for Iris data:} The $2$-layer neural network has $4$ nodes in the input layer, $4$ neurons in the hidden layer and $3$ neurons in the output layer. Thus, the total number of learnable parameters (weights and biases) for this architecture are: ($4 \times 4) + 4 + (3 \times 4) + 3 = 35$. These parameters are passed through a channel corrupted by AWGN with zero mean and increasing  standard deviation (from $0.0003$ to $0.15$). This results in a variation of the Signal-to-Noise Ratio (SNR) from $-8.78$ $dB$ to $45.48$ $dB$. 

\begin{figure}[!h]
	\centering
	    \includegraphics[scale=0.11]{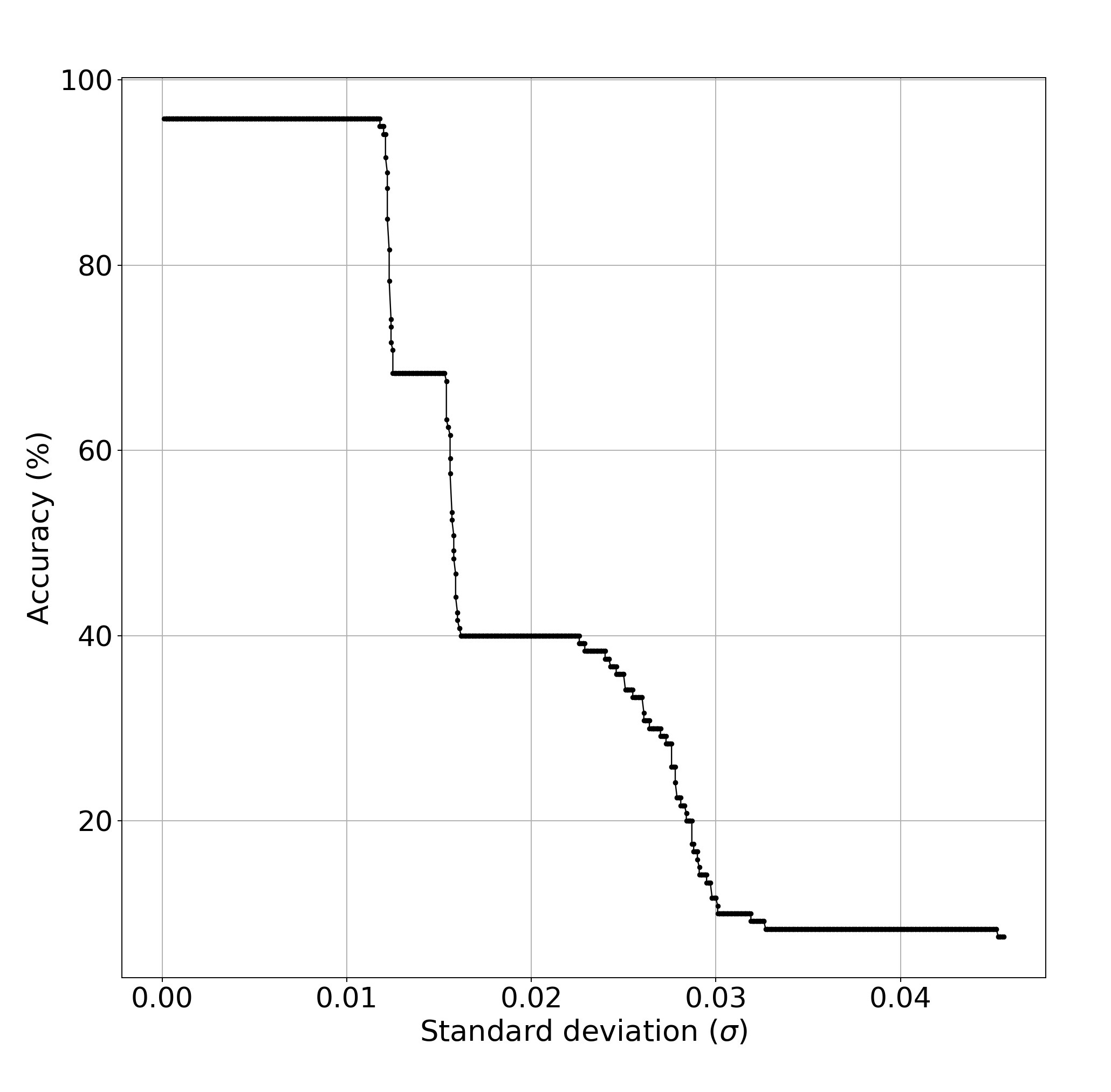}
	    \caption{Parameter noise analysis: Accuracy of single-layer TT-SS algorithm on ChaosNet for Iris data in the presence of AWGN noise with zero mean and increasing standard deviation ($\sigma$).}\label{Fig_ttss_noise_iris}
\end{figure}

\begin{figure}[!h]
    \centering
		\includegraphics[scale=0.11]{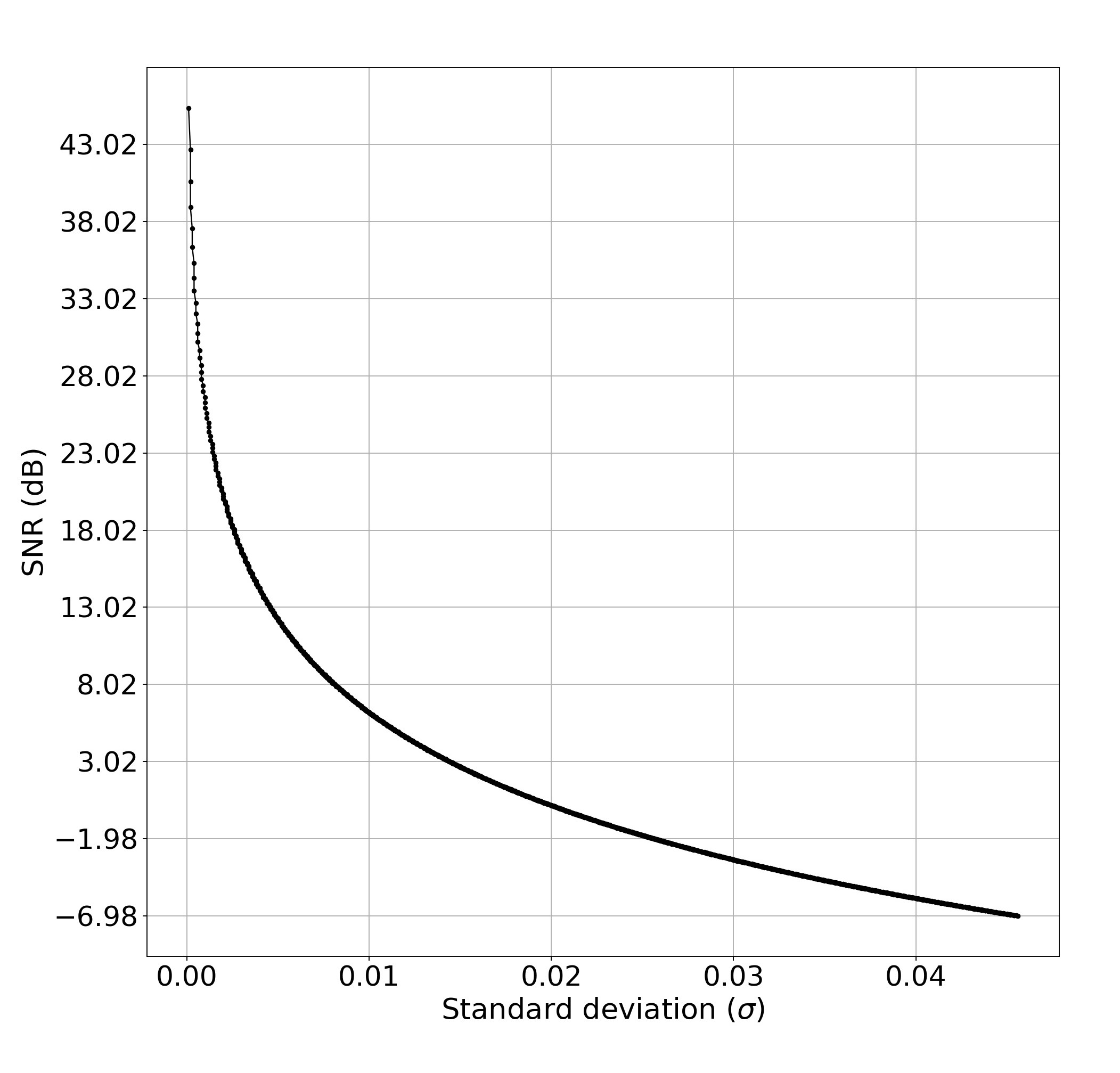}
		\caption{SNR vs. standard deviation ($\sigma$) of AWGN corresponding to Figure~\ref{Fig_ttss_noise_iris}.}\label{Fig_ttss_noise_snr_iris}
\end{figure}

\begin{figure}[!h]
    \centering
	    \includegraphics[scale=0.11]{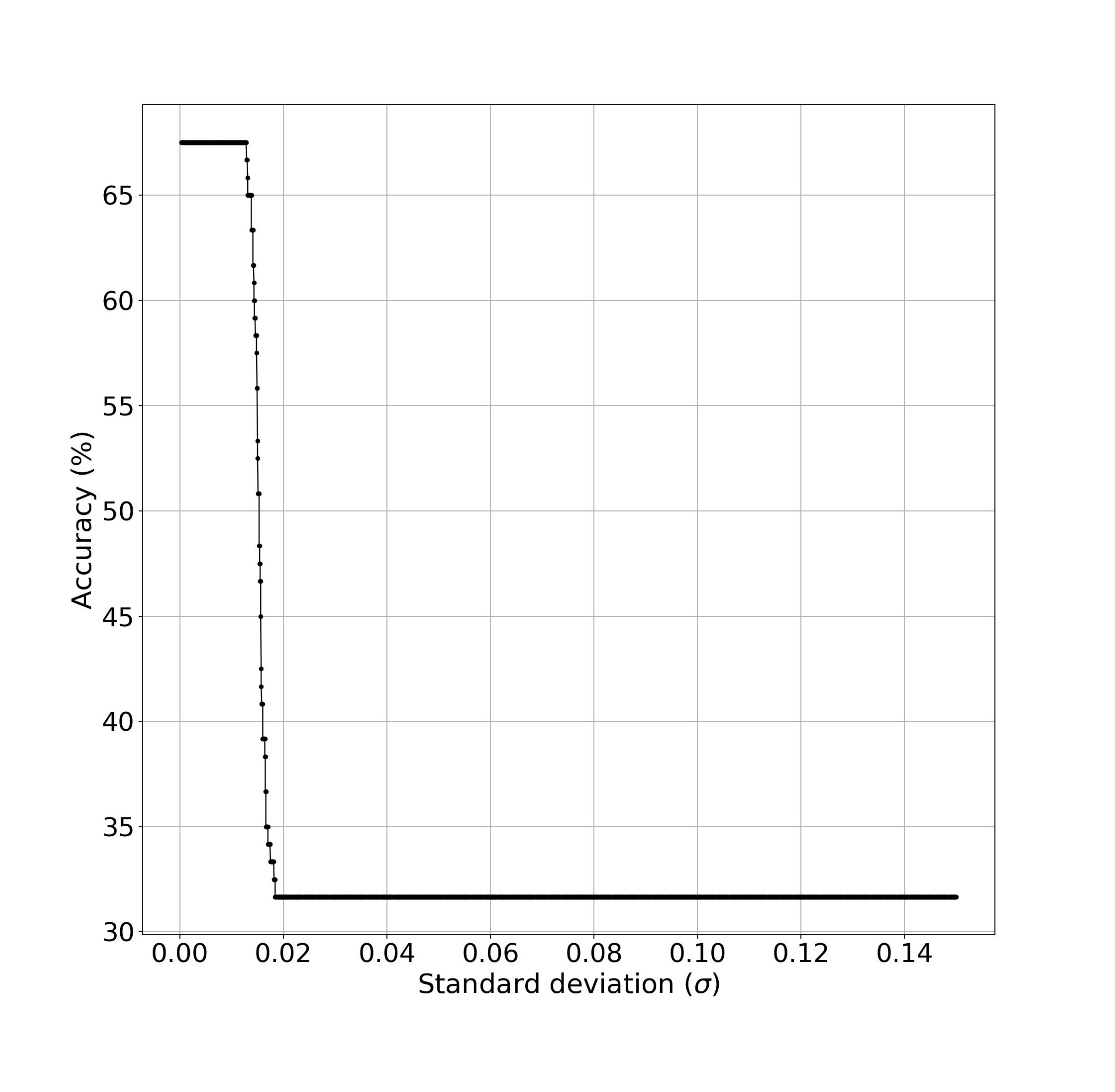}
	    \caption{Parameter noise analysis: Accuracy of 2-layer neural network  for Iris data in the presence of AWGN noise with zero mean and increasing standard deviation ($\sigma$).}\label{Fig_dl_noise_iris}
\end{figure}

\begin{figure}[!h]
    \centering
		\includegraphics[scale=0.11]{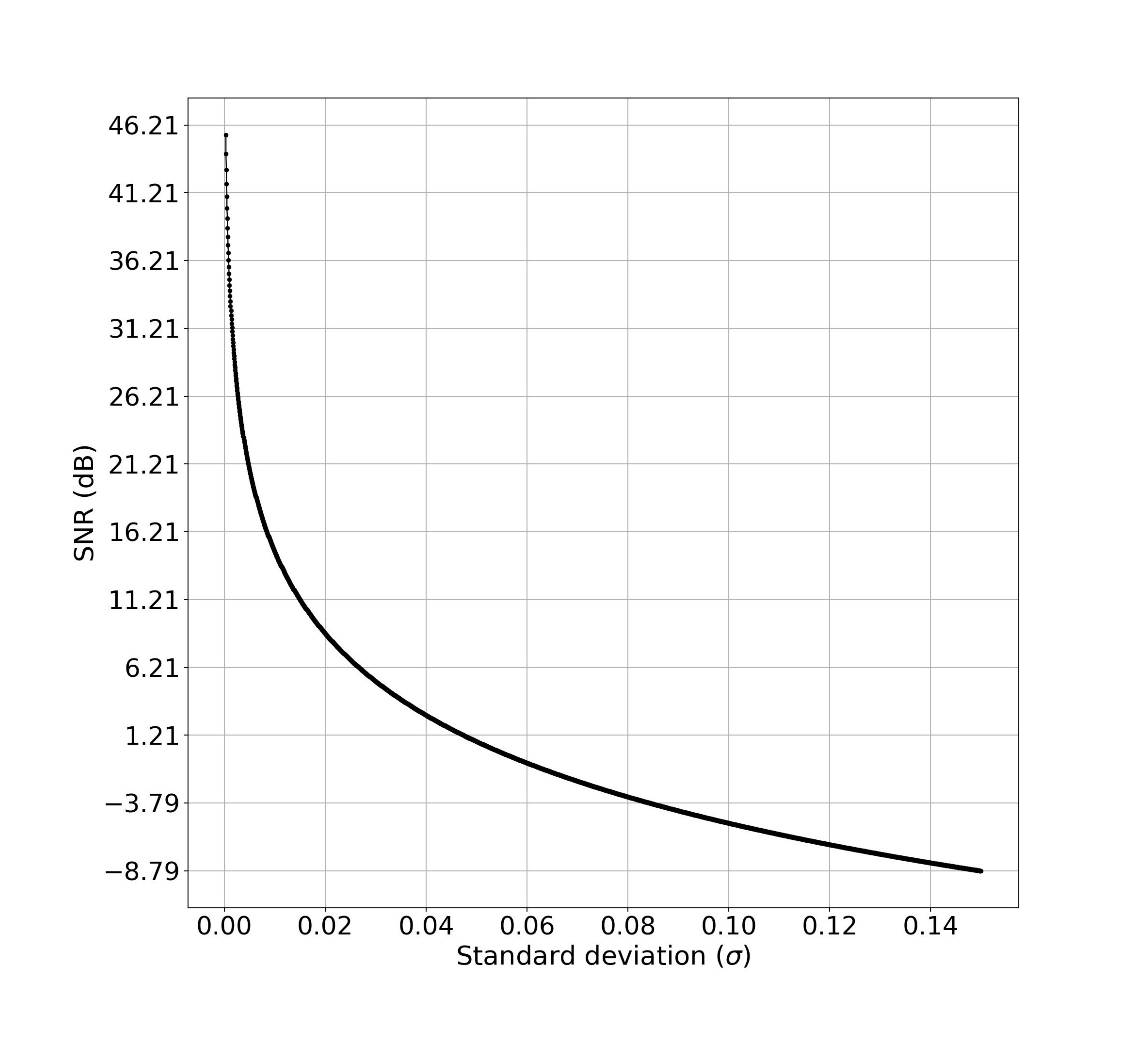}
		\caption{SNR vs. standard deviation ($\sigma$) of AWGN corresponding to Figure~\ref{Fig_dl_noise_iris}.}\label{Fig_dl_noise_snr_iris}
		
\end{figure}

{\it Parameter Noise Analysis:}
 The results corresponding to additive gaussian parameter noise for \verb+ChaosNet+ TT-SS algorithm and $2$-layer neural network are provided in Figure~\ref{Fig_ttss_noise_iris} and Figure~\ref{Fig_dl_noise_iris} respectively. Figure~\ref{Fig_ttss_noise_snr_iris} and Figure~\ref{Fig_dl_noise_snr_iris} depict the variation of SNR (dB) with $\sigma$ of the AWGN for the two algorithms. Firstly, we can observe that the performance of \verb+ChaosNet+ TT-SS algorithm degrades gracefully with increasing $\sigma$, whereas for the $2$-layer neural network there is a sudden and drastic fall in performance (as $\sigma > 0.02$.). A closer observation of the variation of accuracy for different SNRs is provided in Table~\ref{table_snr_acc_iris}. In the case of 1-layer \verb+ChaosNet+ TT-SS method, for the SNR in the range $4.79$ $dB$ to $45.36$ $dB$, the accuracy remains unchanged at $95.83$\%. Whereas for the same SNR, the accuracy for the 2-layer neural network reduced from $67.5$\% to $31.66$\%. This preliminary analysis on parameter noise indicates the better performance of our method when compared with $2$-layer neural network. However, more extensive analysis will be performed for other datasets and other noise scenarios in the near future.
\begin{table}[!h]
\centering
\caption{\label{table_snr_acc_iris} Parameter Noise Analysis for AWGN noise: comparison of accuracies for 1-layer ChaosNet TT-SS method and 2-layer neural network for various SNR ranges.}
\vspace{0.1in}
\begin{tabular}{lccccc}
\hline
\hline
\begin{tabular}[c]{@{}c@{}}SNR (dB)\\ (1-layer TT-SS)\end{tabular} & \begin{tabular}[c]{@{}c@{}}4.79 -- \\ 45.36\end{tabular} & \begin{tabular}[c]{@{}c@{}}4.61 -- \\ 4.76\end{tabular} & \begin{tabular}[c]{@{}c@{}}4.41 -- \\ 4.59\end{tabular} & \begin{tabular}[c]{@{}c@{}}4.27 -- \\ 4.39 \end{tabular} \\ 
\hline
\begin{tabular}[c]{@{}c@{}}Accuracy (\%)\\  (1-layer  TT-SS)\end{tabular} & 95.83 & 95.00 & \begin{tabular}[c]{@{}c@{}}81.66 -- \\ 94.16\end{tabular} & \begin{tabular}[c]{@{}c@{}} 70.83 -- \\ 78.33\end{tabular} \\ 
\hline
\hline
\begin{tabular}[c]{@{}c@{}}SNR (dB)\\ (2-layer NN)\end{tabular} & \begin{tabular}[c]{@{}c@{}}12.56 -- \\ 45.48 \end{tabular} & \begin{tabular}[c]{@{}c@{}}11.94 -- \\ 12.36 \end{tabular} & \begin{tabular}[c]{@{}c@{}}11.57 -- \\ 11.91 \end{tabular} & \begin{tabular}[c]{@{}c@{}}10.65 -- \\ 11.54\end{tabular} \\ 
\hline
\begin{tabular}[c]{@{}c@{}}Accuracy (\%)\\ (2-layer NN)\end{tabular} & 67.50 & 65.00 & \begin{tabular}[c]{@{}c@{}}60.00 -- \\63.33 \end{tabular} & \begin{tabular}[c]{@{}c@{}}40.83 -- \\59.16 \end{tabular} \\ 
\hline
\hline
\end{tabular}
\end{table}
%
%

\newpage
\begin{figure}[!h]
	\centering
	    \includegraphics[scale=0.35]{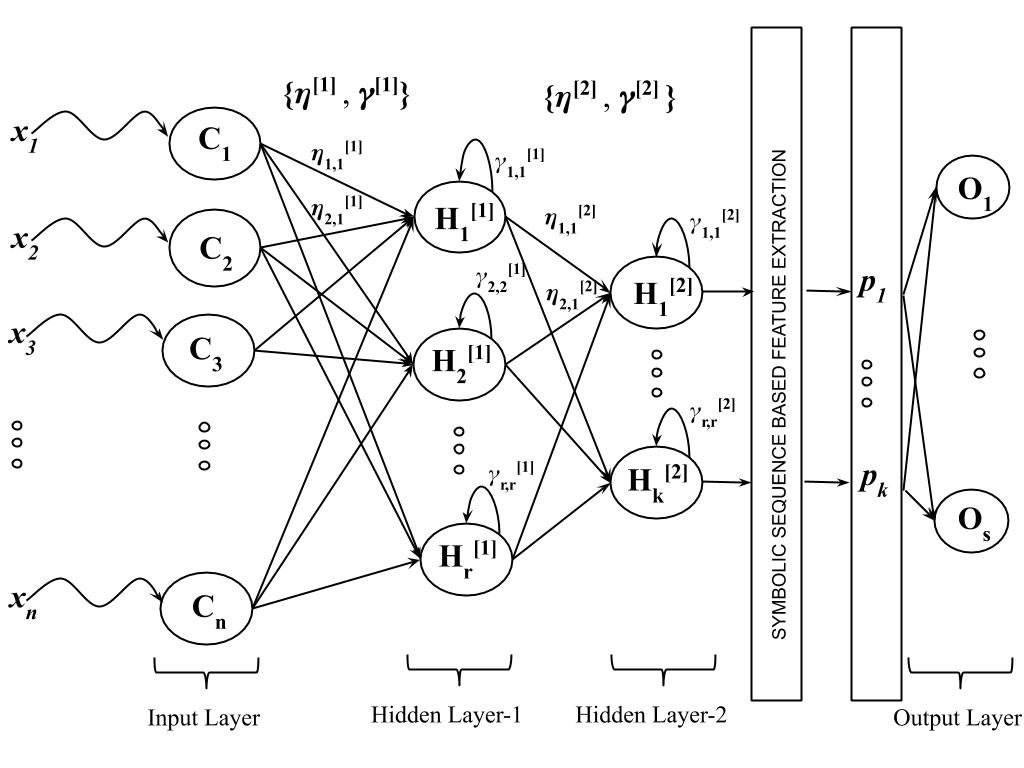}
	
		\caption{Multilayer ChaosNet architecture with two hidden layers and one output layer. Here, the $\eta$s and $\gamma$s are the additional hyperparameters. There is a dimensionality reduction in the representation vectors (if $k<n$).   }\label{Fig_multi_layer_chaosnet}
\end{figure}
\section{Multilayer ChaosNet TT-SS Algorithm}
So far we have discussed  TT-SS algorithm implemented on single-layer \verb+ChaosNet+ architecture. In this section, we investigate the addition of hidden layers to the \verb+ChaosNet+ architecture. A three layer \verb+ChaosNet+ with two hidden layers and one output layer is depicted in Figure~\ref{Fig_multi_layer_chaosnet}. It consists of an input layer with $n$-GLS neurons $\{C_1, C_2, \ldots, C_n\}$, and two hidden layers with GLS neurons $\{H_{1}^{[1]}, H_{2}^{[1]}, \ldots, H_{r}^{[1]}\}$ and $\{H_{1}^{[2]}, H_{2}^{[2]}, \ldots, H_{k}^{[2]}\}$ respectively. Let the neural activity of the input layer GLS neurons be represented by the chaotic trajectories $\bm{A_{1}}$, $\bm{A_{2}}$, \ldots, $\bm{A_{n}}$ with firing times $N_1$, $N_2$, \ldots, $N_n$ respectively and $N_{max} =\max\{N_i\}_{i = 1}^{n}$ denotes the maximum firing time. The $j$-th GLS neuron in the hidden layer has its own intrinsic dynamics starting with an initial neural activity of $q_{j}^{[1]}$ (represented by a self connection with a coupling constant $\gamma$) and potentially connected to every GLS neuron in the input layer (represented by coupling constant $\eta$). 

The coupling coefficient connecting the $i$-th GLS neuron in the input layer to the $j$-th GLS neuron in the first hidden layer of the multilayer \verb+ChaosNet+ architecture is $\eta_{i,j}^{[1]}$. 
The output of the $j$-th GLS neuron of the first hidden layer $(H_{j}^{[1]}(t), j = 1, 2, \ldots, r)$ is as follows:
\begin{equation}
    H_{j}^{[1]}(t)=
    \begin{cases}
      \sum_{i = 1}^{n} \eta_{i,j}^{[1]} A_{i}(t) + \gamma_{j,j}^{[1]} q_{j}^{[1]}, &\ t=0,\\
      \sum_{i = 1}^{n} \eta_{i,j}^{[1]} A_{i}(t) + \gamma_{j,j}^{[1]} T(H_{j}^{[1]}(t-1)), & 0 < t \leq N_{max},
    \end{cases}
  \end{equation}
  
 where $\sum_{i = 1}^{n}\eta_{i,j}^{[1]} + \gamma_{j,j}^{[1]} = 1$, $q_{j}^{[1]} \in (0,1)$ and $\forall$ $t > N_i$, $A_{i}(t) = 0$. The output of the $j$-th neuron of the second hidden layer $(H_{j}^{[2]}(t), j = 1, 2, \ldots, k)$ is as follows:
\begin{equation}
    H_{j}^{[2]}(t)=
    \begin{cases}
      \sum_{i = 1}^{r} \eta_{i,j}^{[2]}  H_{i}^{[1]}(t) + \gamma_{j,j}^{[2]} q_{j}^{[2]}, &  t=0,\\
      \sum_{i = 1}^{r} \eta_{i,j}^{[2]} H_{i}^{[1]}(t) + \gamma_{j,j}^{[2]} T(H_{j}^{[2]}(t-1)), & 0 < t \leq N_{max},
    \end{cases}
  \end{equation}
  
  
  where $\sum_{i = 1}^{r}\eta_{i,j}^{[2]} + \gamma_{j,j}^{[2]} = 1$, $q_{j}^{[2]} \in (0,1)$. In the above equations, $T(\cdot)$ represents the $1$-D chaotic GLS map. From the output of the second hidden layer, the TT-SS features are extracted from the $k$ GLS neurons which are subsequently passed to the output layer for computation and storage of the mean representation vectors corresponding to the $s$ classes. 
  
  In the multilayer \verb+ChaosNet+ TT-SS method, $\eta$s and $\gamma$s are the additional hyperparameters. The above algorithm can be extended in a straightforward manner for more than $2$ hidden layers. 

\begin{figure}[!h]
	\centering
	    \includegraphics[scale=0.031]{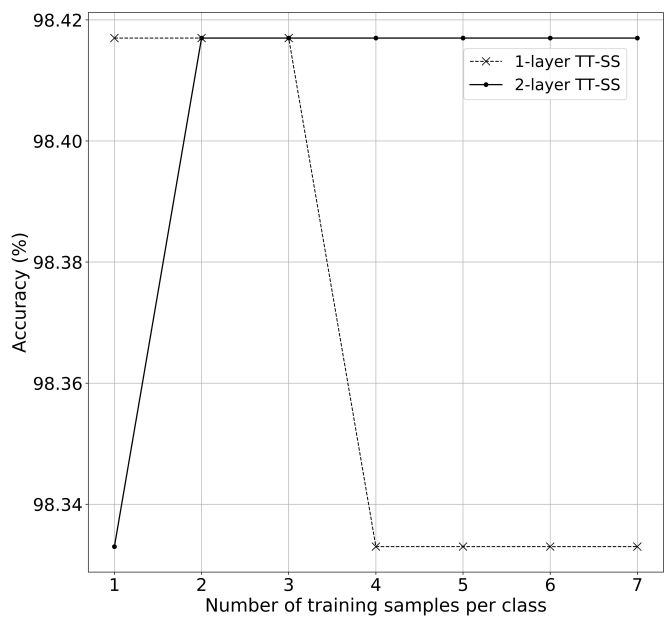}
	
		\caption{Multilayer ChaosNet TT-SS method for Exoplanet dataset. Classification accuracy vs. number of training samples per class for 1-layer TT-SS and 2-layer TT-SS methods.}\label{Fig_comparison_1_layer_2_layer_exoplanet}
	\end{figure}
\subsection{Multilayer ChaosNet TT-SS method for Exoplanet dataset}
We have implemented the multilayer \verb+ChaosNet+ TT-SS method for the Exoplanet dataset with 2 layers (one hidden layer and one output layer). The number of GLS neurons in the input and first hidden layer are $n=45$ and $r=23$ respectively. The output layer consists of $s=3$ nodes as it is a $3$-class classification problem ({\it Mesoplanets}, {\it Psychroplanets} and {\it Non-Habitable}). Every neuron in the first hidden layer is connected to only two neurons in the input layer (except the last neuron which is connected to only one neuron in the input layer). The output of the GLS neurons in the hidden layer is given by: 
\begin{equation}
     H_{j}^{[1]}(t)=
    \begin{cases}
       \eta_{2j-1,j}^{[1]} A_{2j-1}(t)+\eta_{2j,j}^{[1]} A_{2j}(t) +\\ \gamma_{j,j}^{[1]} q_{j}^{[1]}, &\ t=0,\\
      \eta_{2j-1,j}^{[1]} A_{2j-1}(t)+\eta_{2j,j}^{[1]} A_{2j}(t) +\\ \gamma_{j,j}^{[1]} T(H_{j}^{[1]}(t-1)), & 0 < t \leq N_{max},
    \end{cases}
    \label{eq:multi_layer_equation}
\end{equation}

for $j = 1, 2, \ldots, 22$. To the last neuron of first hidden layer the input value is passed as such. The hyperparameters used in the classification task are: $\eta = 0.4995$ and  $\gamma = 0.001$, $q_j^{[1]} = 0.56$; initial neural activity $q$, internal discrimination threshold ($b$), type of GLS neuron used and $\epsilon$ chosen for Exoplanet classification task are the same as in Table~\ref{table_parameters_ttss} (fourth row).

From Figure~\ref{Fig_comparison_1_layer_2_layer_exoplanet}, $2$-layer \verb+ChaosNet+ TT-SS method has slightly improved the accuracy of Exoplanet classification task over that of $1$-layer \verb+ChaosNet+ TT-SS method for four and higher number of training samples per class. There is a $50\%$ reduction in the dimensionality of the representation vectors (at the cost of increase in the number of hyperparameters). While these preliminary results are encouraging, more extensive testing of multilayer \verb+ChaosNet+ TT-SS method with fully connected layers (and more than one hidden layer) need to be performed in the future. 

\section{\label{sec:Conclusions} Conclusions and Future Research Directions}
State-of-the-art performance on classification tasks reported by algorithms in literature are typically for $80\%-20\%$ or $90\% - 10\%$ training-testing split of the data-sets. The performance of these algorithms will dip considerably as the number of training samples reduces. \verb+ChaosNet+ demonstrates ($1$-layer TT-SS method) consistently good (and reasonable) performance accuracy in the low training sample regime. Even with as low as $7$ (or fewer) training samples/class (which accounts for less than 0.05\% of the total available data), \verb+ChaosNet+ yields performance accuracies in the range $73.89 \% - 98.33 \%$.

Future work includes determining optimal hyperparameter settings to further improve accuracy, testing on more datasets and classification tasks, extension to predictive modelling and incorporating robustness into GLS neurons to external noise. Multilayer \verb+ChaosNet+ architecture presents a number of avenues for further research such as determining optimal number of layers, type of coupling (unidirectional and bidirectional) between layers, homogeneous and heterogeneous layers (successive layers can have neurons with different 1D chaotic maps), coupled map lattices, use of 2D and higher dimensional chaotic maps and even flows in the architecture and exploring properties of chaotic synchronization in such networks. 

Highly desirable features such as Shannon optimal lossless compression, computation of logical operations (XOR, AND etc.), universal approximation property and topological transitivity -- all thanks to the chaotic nature of GLS neurons --  makes \verb+ChaosNet+ a potentially attractive ANN architecture for diverse applications (from memory encoding for storage and retrieval purposes to classification tasks).  We expect design and implementation of novel learning algorithms on the \verb+ChaosNet+ architecture in the near future that can efficiently exploit these wonderful properties of chaotic GLS neurons. 

The code for the proposed \verb+ChaosNet+ architecture (TT-SS method) is available at~\url{https://github.com/HarikrishnanNB/ChaosNet}.

\section*{Acknowledgment}
Harikrishnan N. B. thanks ``The University of Trans-Disciplinary Health Sciences and Technology (TDU)'' for permitting this research as part of the PhD programme. Aditi Kathpalia is grateful to the Manipal Academy of Higher Education for permitting this research as a part of the PhD programme. The authors gratefully acknowledge the financial support of Tata Trusts. N. N. dedicates this work to late Prof. Prabhakar G Vaidya who initiated him to the fascinating field of Chaos Theory.
%
%


%
\newpage
\section*{Appendix}
\subsection*{(I). Algorithms for the 1-layer TT-SS Method implemented on ChaosNet}
\begin{figure}[!h]
\centerline{ \includegraphics[scale=0.11]{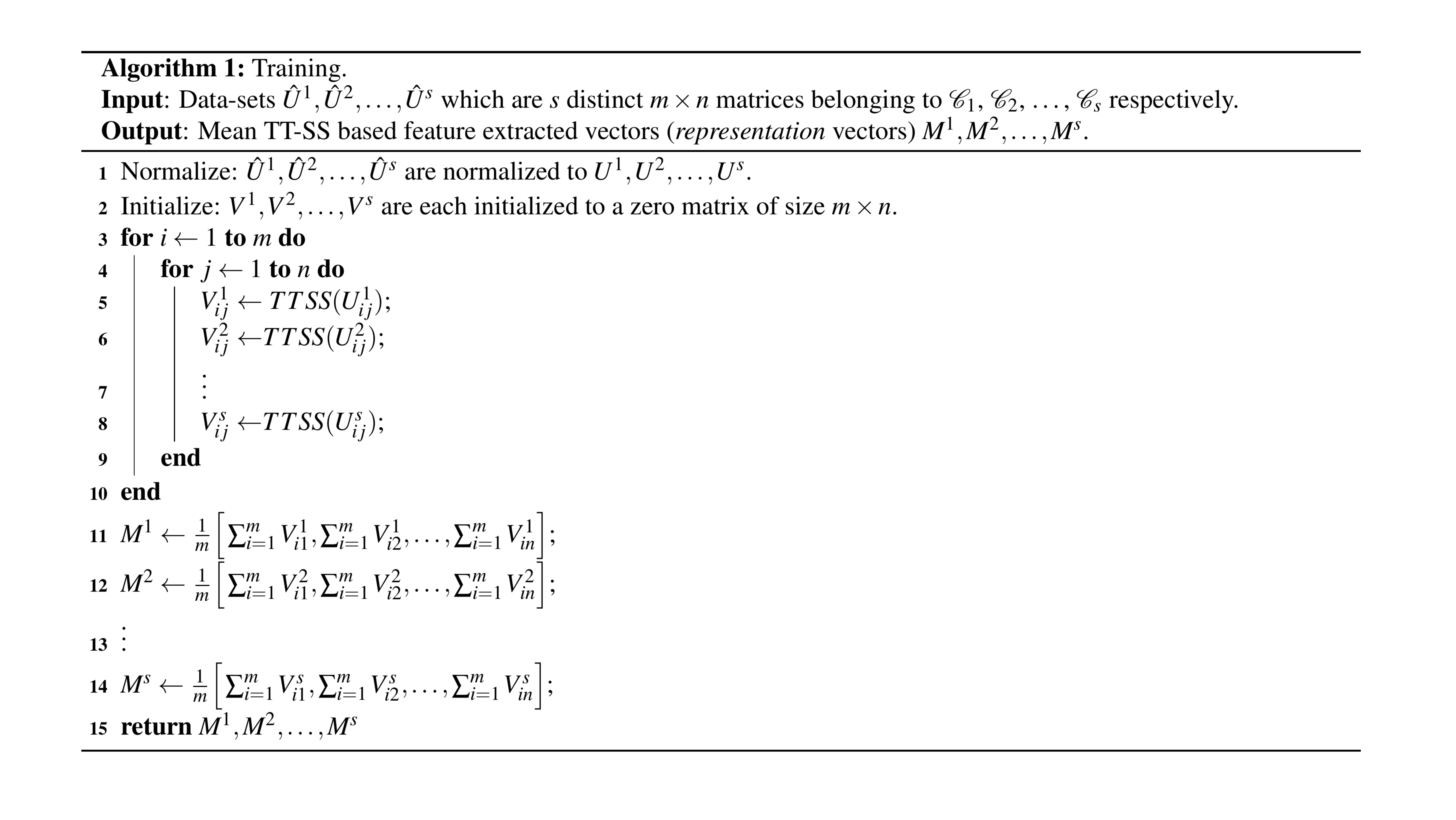}}
\label{alg:TT}
\end{figure}
\begin{figure}[!h]
\centerline{ \includegraphics[scale=0.11]{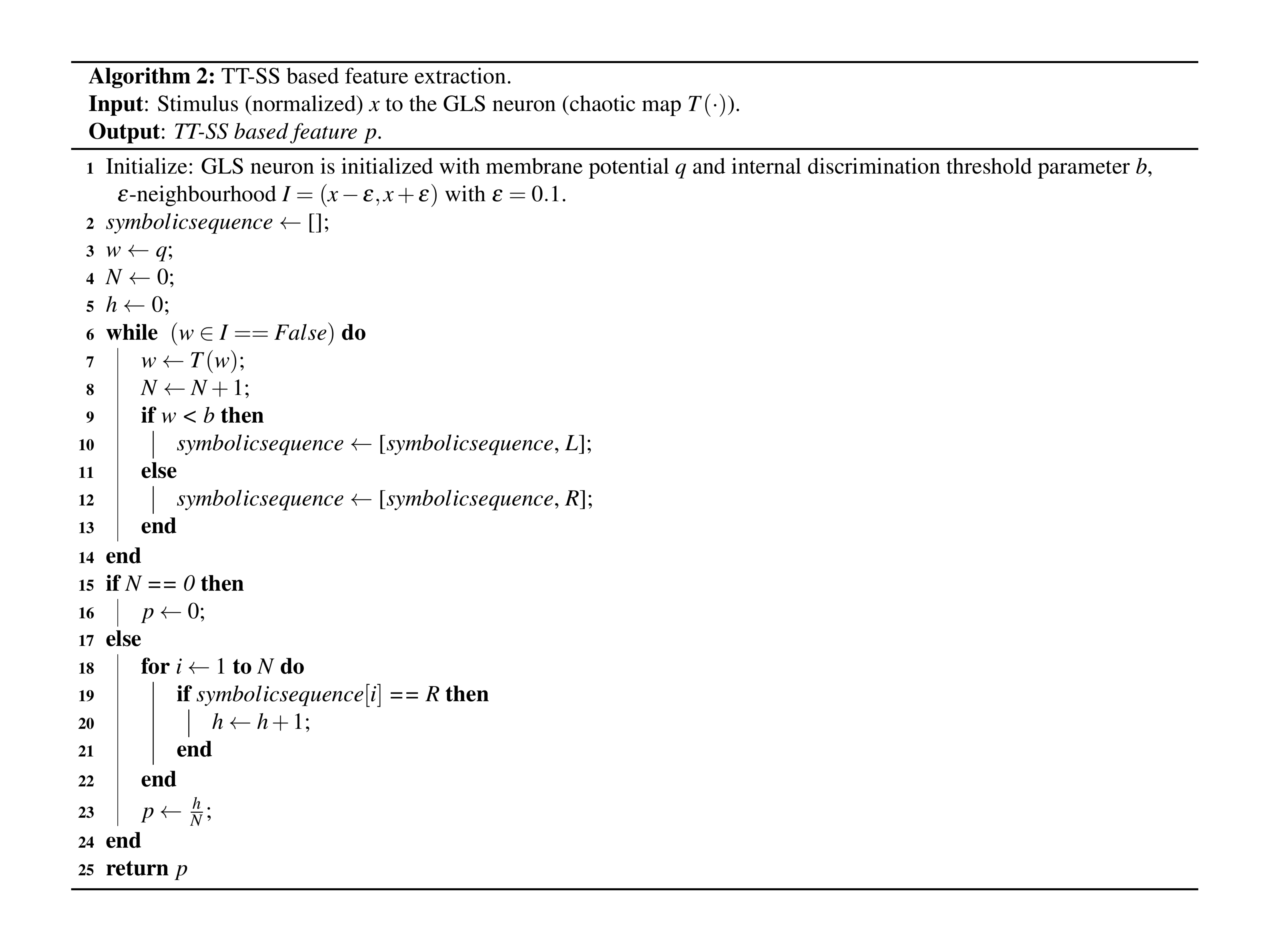}}
\label{alg:train}
\end{figure}
\begin{figure}[!h]
\centerline{ \includegraphics[scale=0.11]{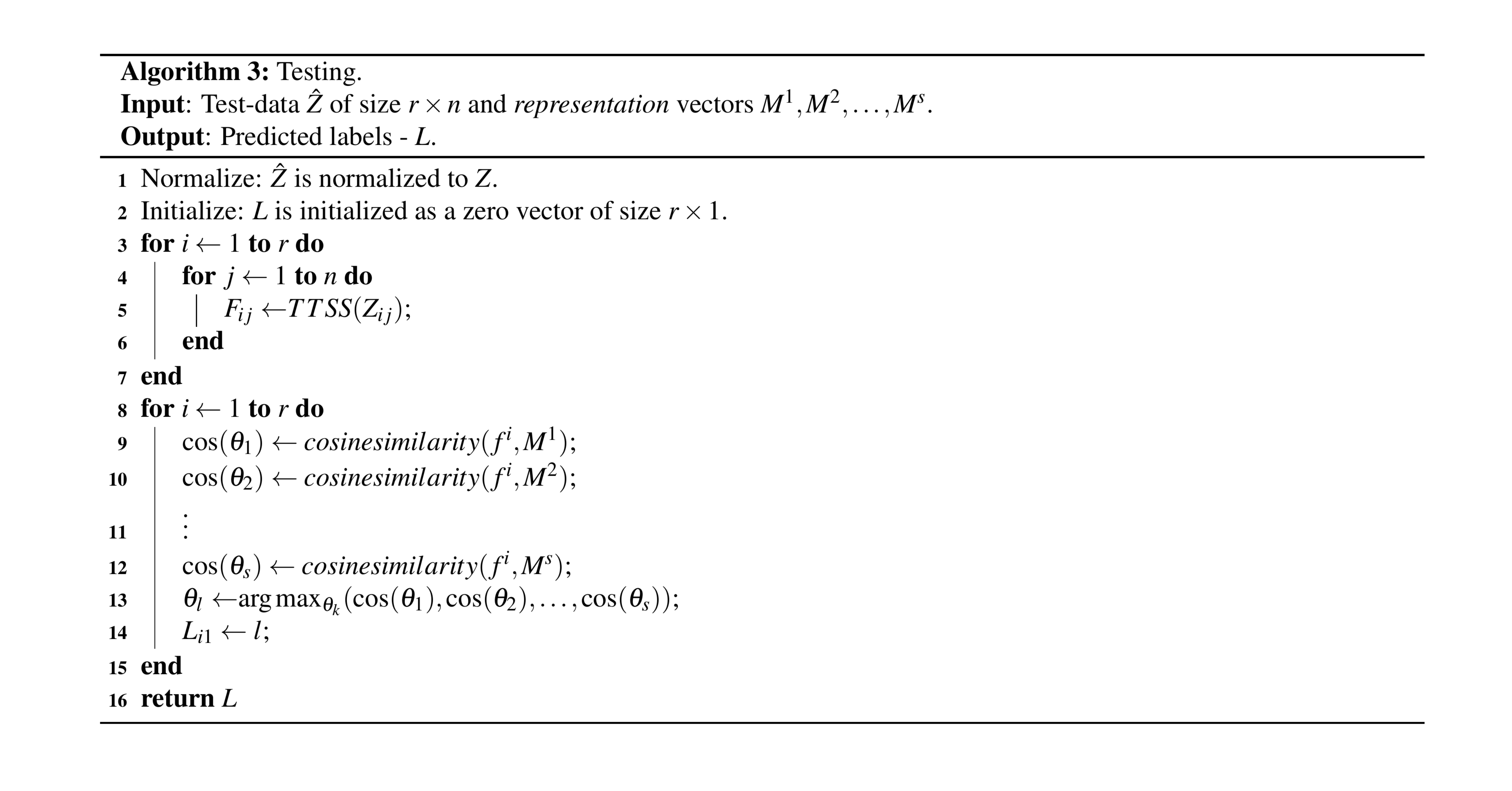}}
\label{alg:test}
\centerline{ \includegraphics[scale=0.11]{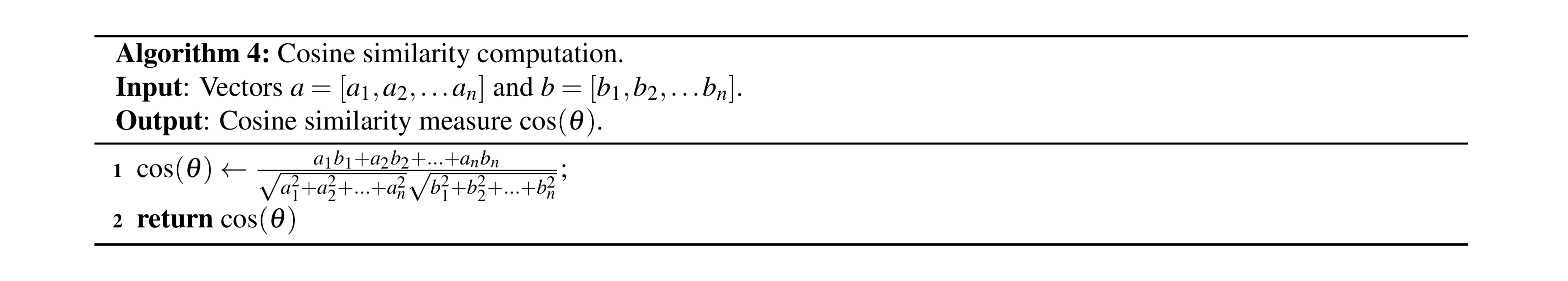}}
\label{alg:cosinesimilarity}
\end{figure}

\newpage
\clearpage
\subsection*{(II). Hyperparameters used by machine learning algorithms in {\it scikit-learn} and {\it Keras}}
The following table lists the hyperparameters that we have used for the machine learning algorithms for generating the results in the main manuscript.

\begin{table}[!h]
\renewcommand{\arraystretch}{1.3}
\centering
\caption{Hyperparameters settings for the ML algorithms used in the main manuscript.}
\label{ML-parameters}
\scalebox{0.85}{
\begin{tabular}{|c|c|c|}
\hline
Datasets & Algorithms & Hyperparameters\\\hline
MNIST & ~~~~~~~~~~ & class\_weight=None, criterion='gini', max\_depth=None,  max\_features=None,\\
KDDCup'99 &~~~~~~~~ &       max\_leaf\_nodes=None, min\_impurity\_decrease=0.0, min\_impurity\_split=None,\\
Iris  &Decision Tree&       min\_samples\_leaf=1, min\_samples\_split=2, min\_weight\_fraction\_leaf=0.0, presort=False,   \\
Exoplanet &~~~~~~~~ &  random\_state=1234, splitter='best'            \\\hline
MNIST & ~~~~~~~~~~ &            C=1.0, cache\_size=200, class\_weight=None, coef0=0.0,\\
KDDCup'99 &~~~~~~~~ &            decision\_function\_shape='ovr', degree=3, gamma='auto', kernel='rbf',\\
Iris  &SVM &              max\_iter=-1, probability=False, random\_state=None, shrinking=True,\\
Exoplanet &~~~~~~~~ &             tol=0.001, verbose=False\\\hline
MNIST & KNN &            algorithm='auto', leaf\_size=30, metric='minkowski',\\
~~~~~~~&~~~~~~~~~~&    metric\_params=None, n\_jobs=-1, n\_neighbors= 5, p=2, weights='uniform'\\ \hline

KDDCup'99 & KNN &            algorithm='auto', leaf\_size=30, metric='minkowski',\\
~~~~~~~~~~~&~~~~~~~~~~&    metric\_params=None, n\_jobs=-1, n\_neighbors= 9, p=2, weights='uniform'\\ \hline

Iris  & KNN &             algorithm='auto', leaf\_size=30, metric='minkowski',\\
Exoplanet&~~~~~~~~~~~&    metric\_params=None, n\_jobs=-1, n\_neighbors= 3, p=2, weights='uniform'\\ \hline
MNIST &~~~~~~~~~~~~~~~~   & number of neurons in 1st hidden layer = 784, activation='relu',\\
~~~~~~~~~~~& 2-layer neural network & number of neurons in output layer = 10, activation = 'softmax',\\ ~~~~~~~~~~&~~~~~~~~~~~~&   loss='categorical\_crossentropy', optimizer='adam'\\ \hline
KDDCup'99 &   & number of neurons in 1st hidden layer = 41, activation='relu',\\
~~~~~~~~~~~& 2-layer neural network & number of neurons in output layer = 9, activation = 'softmax',\\ ~~~~~~~~~~&~~~~~~~~~~~~&   loss='categorical\_crossentropy', optimizer='adam'\\ \hline
Iris &  ~~~~~~~~~~~~~~~~~ & number of neurons in 1st hidden layer = 4, activation='relu',\\
~~~~~~~~~~~& 2-layer neural network & number of neurons in output layer = 3, activation = 'softmax',\\ ~~~~~~~~~~&~~~~~~~~~~~~&   loss='categorical\_crossentropy', optimizer='adam'\\ \hline
Exoplanet &  ~~~~~~~~~~~~~~~~~ & number of neurons in 1st hidden layer = 45, activation='relu',\\
~~~~~~~~~~~& 2-layer neural network & number of neurons in output layer = 3, activation = 'softmax',\\ ~~~~~~~~~~&~~~~~~~~~~~~&   loss='categorical\_crossentropy', optimizer='adam'\\ \hline
Exoplanet  &~~~~~~~~~~~~~~~~~~~~~& number of neurons in 1st hidden layer = 42, activation='relu',\\
with no & 2-layer neural network & number of neurons in output layer = 3, activation = 'softmax',\\ 
surface temperature &~~~~~~~~~~~~&   loss='categorical\_crossentropy', optimizer='adam'\\ \hline
Exoplanet  &~~~~~~~~~~~~~~~~~~~~~& number of neurons in 1st hidden layer = 6, activation='relu',\\
with 6 & 2-layer neural network & number of neurons in output layer = 3, activation = 'softmax',\\ 
restricted features &~~~~~~~~~~~~&   loss='categorical\_crossentropy', optimizer='adam'\\ \hline
\end{tabular}
}
\end{table}
\end{document}